\documentclass[journal]{IEEEtran}
\ifCLASSOPTIONcompsoc
  \usepackage[nocompress]{cite}
\else
  \usepackage{cite}
\fi
\ifCLASSINFOpdf
\else
\fi
\usepackage{amsmath,amsfonts}
\usepackage{algorithmic}
\usepackage{array}
\usepackage{textcomp}
\usepackage{stfloats}
\usepackage{url}
\usepackage{verbatim}
\usepackage{graphicx}
\usepackage{amssymb}
\usepackage{bm}
\usepackage{subcaption}
\usepackage{multirow}
\usepackage{color}
\usepackage{cite}
\usepackage{tabularx}
\usepackage[table]{xcolor}

\usepackage[switch]{lineno}
\usepackage{hyperref}
\usepackage{bbding}
\usepackage{booktabs}
\usepackage{makecell}

\def\thetaset{\bm{\theta}}
\def\Thetaset{\bm{\Theta}}
\def\P{\mathbf{P} }
\def\Xt{\widetilde{\mathbf{X}}}
\def\Yt{\widetilde{\mathbf{Y}}}
\def\xt{\widetilde{\mathbf{x}}}
\def\yt{\widetilde{\mathbf{y}}}
\def\xhat{\hat{\mathbf{x}}}
\def\yhat{\hat{\mathbf{y}}}

\def\N{\mathcal{N}}
\def\NxX{\N_{\x_i,K_1}^{\X}}
\def\NXX{\N_{\X,K_1}^{\X}}

\def\NxXl{\N_{\x_i,K_2}^{\X_l}}

\def\NxY{\N_{\x_i,K}^{\Y}}
\def\NXY{\N_{\X,K}^{\Y_l}}
\def\NxXhat{\N_{\xhat_i,K}^{\Xhat_l}}

\def\p{\mathbf{p}}
\def\x{\mathbf{x}}
\def\y{\mathbf{y}}
\def\s{\mathbf{s}}
\def\e{\mathbf{e}}
\def\r{\mathbf{r}}
\def\v{\mathbf{v}}
\def\X{\mathbf{X}}
\def\Y{\mathbf{Y}}
\def\I{\mathbf{I}}
\def\E{\mathbf{E}}
\def\S{\mathbf{S}}
\def\R{\mathbb{R}}
\def\V{\mathbf{V}}
\def\D{\mathbf{D}}
\def\W{\mathbf{W}}
\def\L{\mathbf{L}}
\def\u{\mathbf{u}}
\def\t{\mathbf{t}}
\def\v{\mathbf{v}}
\def\C{\mathbf{C}}
\def\c{\mathbf{c}}
\def\G{\mathcal{G}}
\def\b{\mathbf{b}}
\def\M{\mathbb{M}}
\def\A{\mathbb{A}}

\def\Tr{\mathrm{Tr}}

\definecolor{lightgreen}{RGB}{226, 245, 226}
\definecolor{softblack}{RGB}{85, 85, 85}

\begin{document}

\title{OuroReward: Sequential Reward Scheduling for Reinforcement Learning in Text-to-3D Generation}

\markboth{Journal of \LaTeX\ Class Files,~Vol.~14, No.~8, August~2015}%
{Shell \MakeLowercase{\textit{et al.}}: Bare Demo of IEEEtran.cls for Computer Society Journals}


\author{Bingyang Cui$^{1}$ \quad Yujie Zhang$^{2}$ \quad Yiling Xu$^{1}$~\IEEEmembership{Member,~IEEE} \quad Yunfeng Guan$^{1}$\\
$^{1}$ Shanghai Jiao Tong University\quad \textsuperscript{2} JD.COM  \\
{\tt\small $^{1}$\{ccccby0813, yl.xu, yfguan69\}@sjtu.edu.cn , $^{2}$zhangyujie.1998@jd.com}
}

\IEEEtitleabstractindextext{%

\begin{abstract}
Reinforcement learning (RL) for Text-to-3D (T23D) generation requires optimization across multiple quality dimensions such as semantic alignment and texture clarity. Existing methods typically optimize these dimensions simultaneously through multiple reward aggregation, without explicitly modeling inter-dimension dependencies. This can cause imbalanced optimization and persistent interference among conflicting dimensions. To address this limitation, we propose \textbf{OuroReward}, an interference-aware sequential reward scheduling strategy for T23D RL. OuroReward first estimates pairwise dependencies among dimensions and constructs a cyclic optimization path that minimizes cumulative interference. By incorporating the \textit{tail-to-head} dependency, the cycle captures global compatibility across the entire schedule. Then, OuroReward converts the cycle into a one-pass sequence, and starts optimization from the dimension with the lowest aggregate interference. Rather than assigning a fixed optimization budget to each dimension-wise reward, training adaptively determines when to advance to the next reward according to the remaining optimization headroom of the current one. We further introduce \textbf{AdaSelect}, an adaptive prompt selection strategy that identifies reliable and informative prompts aligned with the model's current capability. By focusing policy updates on these prompts, AdaSelect effectively improves training stability. Extensive experiments across different T23D models and RL algorithms demonstrate that our framework consistently improves generation quality across multiple dimensions.
\end{abstract}

}

\maketitle
\IEEEdisplaynontitleabstractindextext

%
\IEEEpeerreviewmaketitle

\section{Introduction}\label{sec:introduction}

\begin{figure}[!t]
    \centering
    \begin{subfigure}{\linewidth}
        \centering
        \includegraphics[width=\linewidth]{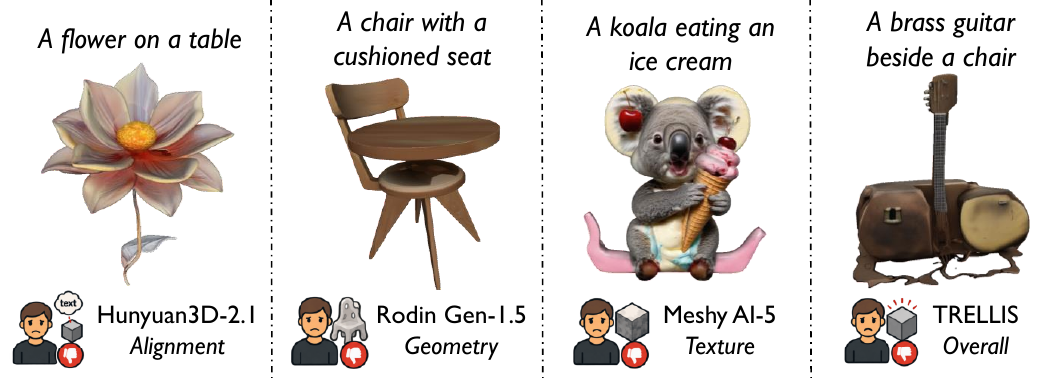}
        \caption{}
        \label{fig:motivation_generation}
    \end{subfigure}
    \hfill
    \begin{subfigure}{\linewidth}
        \centering
        \includegraphics[width=\linewidth]{image/motivation_method.png}
        \caption{}
        \label{fig:motivation_method}
    \end{subfigure}
    \caption{(a) Representative failure cases produced by different T23D generation models. (b) Comparison of gradient discrepancy and reward-change conflict ratio between a conflicting dimension pair (\textit{G}$\leftrightarrow$\textit{T}) and a compatible dimension pair (\textit{A}$\leftrightarrow$\textit{O}) under the default reward-averaging strategy. \textit{A}, \textit{G}, \textit{T}, and \textit{O} denote four quality dimensions including \textit{Alignment}, \textit{Geometry}, \textit{Texture}, and \textit{Overall}, respectively.}
    \label{fig:motivation_illustration}
    \vspace{-0.4cm}
\end{figure}

\IEEEPARstart{T}{ext-to-3D} (T23D) generation has attracted increasing attention for its ability to synthesize 3D content from textual prompts \cite{hunyuan3d21,chen2025meshanything,zhao2025deepmesh,tripo,meshy,xiang2025trellis}. Existing T23D models can be broadly categorized into autoregressive-based \cite{xu2026meshweaver,wang2026shapellm,tang2026we,han2026var,yi2026progressive} and diffusion-based approaches \cite{di2025hyper,qian2026gs,wu2026animateanymesh++}, among which diffusion-based models have demonstrated strong capability in modeling complex 3D representations. Nevertheless, as illustrated in Fig.~\ref{fig:motivation_generation}, current T23D models still suffer from quality limitations across multiple dimensions, such as semantic misalignment, geometry degradation, and visually unappealing textures. To further improve generation quality, recent studies have introduced reinforcement learning (RL) into generative models, leveraging reward signals to guide the optimization direction \cite{shao2024deepseekmath,liu2026flowgrpo,zheng2026diffusionnft}. Considering the complexity of 3D patterns, T23D generation requires optimization across multiple quality dimensions, where each dimension is represented by its corresponding reward signal.

Existing multi-dimensional RL methods generally aggregate dimension-wise rewards into a scalar objective and optimize simultaneously within each policy update. However, such joint optimization can bias policy updates toward readily optimizable dimensions, causing them to dominate training and ultimately leading to imbalanced performance across dimensions. To alleviate this issue, FocalReward \cite{huang2026focal} dynamically reweights rewards according to their saturation levels, while other methods adopt multi-task learning formulations to reconcile reward-specific gradients and obtain more compatible update directions \cite{sener2018multi,yu2020gradient,lee2024parrot,jiang2026t2i}. Nevertheless, reward reweighting and gradient reconciliation still operate within the same joint policy update and therefore cannot fundamentally eliminate their intrinsic optimization bias. This motivates sequential reward scheduling, \textit{i.e.}, decoupling competing rewards into separate optimization stages rather than balancing them simultaneously within each policy update.

Effective sequential dimension-wise reward scheduling further requires an appropriate reward order, because different dimension pairs exhibit substantially different degrees of \textit{inter-dimension interference}. To reveal this heterogeneity, we analyze the relationship between pairwise gradient discrepancies and reward-change conflict ratios under the default reward-averaging setting, where the latter denotes the proportion of validation samples for which two rewards change in opposite directions between consecutive training steps. As illustrated in Fig.~\ref{fig:motivation_method}, dimension pairs with larger gradient discrepancies exhibit increasing conflict ratios as training progresses, whereas pairs with smaller discrepancies show decreasing conflict ratios. These observations indicate that inter-dimension dependencies are highly non-uniform across different pairs. Therefore, simply separating dimension-wise reward into sequential stages is insufficient; the reward order should explicitly account for such dependencies to reduce interference between consecutive stages and limit its accumulation over the full optimization sequence.

Motivated by these findings, we propose \textbf{OuroReward}, an interference-aware sequential reward scheduling strategy for multi-dimensional T23D RL. OuroReward first estimates pairwise dimension dependencies using gradient cosine similarity and reward-change correlation during a lightweight warm-up stage. Based on these dependencies, it constructs a cyclic path that maximizes cumulative compatibility across adjacent dimensions. By incorporating the \textit{tail-to-head} dependency, the cycle evaluates compatibility over the entire path rather than only between locally adjacent dimensions. Since starting from a highly conflicting dimension may propagate interference to subsequent stages, OuroReward selects the dimension with the highest overall compatibility as the starting point and unfolds the cycle into a one-pass optimization sequence. During training, it adaptively transitions the reward objective along this sequence according to the optimization headroom of the current dimension, avoiding excessive optimization of any single dimension. In this way, OuroReward aligns sequential optimization with pairwise dependencies, reducing cumulative interference while alleviating the dimension dominance inherent in simultaneous multi-dimensional optimization.

Moreover, the effectiveness of T23D RL strongly depends on whether training prompts are well matched to the model's current capability. Prompts that are too easy may provide limited learning signals, while overly difficult prompts can lead to noisy or unstable optimization. As the model evolves throughout training, the set of informative prompts should therefore change accordingly. To address this issue, we propose \textbf{AdaSelect}, an adaptive prompt selection strategy that dynamically identifies prompts aligned with the model's current capability. AdaSelect evaluates candidate prompts based on their capability compatibility and rollout characteristics, and prioritizes those that provide informative and reliable learning signals at each training stage. By focusing policy updates on capability-matched prompts, AdaSelect improves training stability and effectiveness.

Our contributions can be summarized as threefold:
\begin{itemize}
    \item We propose \textbf{OuroReward}, a novel Ouroboros-inspired sequential reward scheduling strategy for multi-dimensional T23D RL. OuroReward constructs a cyclic dependency path by maximizing cumulative compatibility across adjacent dimensions.  It then selects the most compatible dimension as the starting point and performs one-pass sequential optimization with headroom-aware reward transitions, thereby reducing negative transfer across quality dimensions.
    
    \item We propose \textbf{AdaSelect}, an adaptive prompt selection strategy for improving the stability of T23D RL. AdaSelect identifies informative prompts according to the model's current capability, rollout discriminability, optimization compatibility, and selection history. By prioritizing prompts that yield reliable learning signals, AdaSelect improves training stability.
    
    \item Extensive experiments across multiple T23D models and RL algorithms demonstrate that the proposed framework consistently improves generation quality across different quality dimensions, and substantially enhances the training stability of T23D RL.
\end{itemize}

\section{Related Work}\label{sec:related_work}

\subsection{T23D Generative Models}

T23D generation aims to synthesize 3D content from textual prompts and has attracted substantial attention for enabling scalable and accessible 3D content creation. Following existing surveys \cite{tang2025recentsurvey,jiang2024survey,hunyuan3d21}, current approaches can be broadly categorized into two paradigms: autoregressive-based and diffusion-based generation. Autoregressive-based methods discretize 3D content into token sequences and generate them sequentially, whereas diffusion-based methods progressively recover 3D representations from noise through iterative denoising. Compared with autoregressive-based formulations, diffusion-based approaches are more suitable for T23D generation, offering stronger modeling capability and higher generation fidelity. More recently, diffusion-based models incorporating flow matching have further improved sampling efficiency while maintaining high generation quality, making them particularly suitable for efficient T23D generation and subsequent RL optimization.

\textbf{Autoregressive-based Generation.} 
These models typically discretize 3D content into token sequences and generate them sequentially. Early studies such as MeshGPT \cite{siddiqui2024meshgpt} and MeshAnything \cite{chen2025meshanything} establish this paradigm for triangle and artist-grade mesh generation. Building on this formulation, LLaMA-Mesh \cite{wang2024llama} and DeepMesh \cite{zhao2025deepmesh} further extend autoregressive 3D generation toward large language models (LLMs) and preference optimization. More recent works focus on improving the generation efficiency and structural fidelity. FACE \cite{wang2026face} introduces a face-level representation to substantially shorten mesh token sequences, MeshRipple \cite{lin2026meshripple} incorporates topology-aware sequential modeling to better preserve local connectivity, and MeshWeaver \cite{xu2026meshweaver} leverages geometry-aware guidance for more accurate surface generation. FlashMesh \cite{shen2026flashmesh} further accelerates inference through structured speculative decoding. In parallel, ShapeLLMOmni \cite{wang2026shapellm} develops a unified multimodal LLM for 3D generation and understanding by discretizing 3D shapes with a 3D VQVAE. Despite these advances,autoregressive-based models are still constrained by discrete tokenization and sequential decoding. Discretizing complex 3D geometry may compromise fine-grained geometric and topological details, while the growing sequence length substantially increases decoding cost for complex 3D assets. Consequently, autoregressive-based models remain less scalable for efficient high-fidelity 3D generation.

\textbf{Diffusion-based Generation.}
These models can be broadly divided into optimization-based and feedforward-based approaches. Optimization-based models typically leverage pre-trained multimodal networks to optimize 3D models. DreamFusion \cite{dreamfusion} establishes this paradigm by optimizing Neural Radiance Fields (NeRF) using Score Distillation Sampling (SDS). Subsequent works \cite{di2025hyper,ji2026flux3d,qian2026gs} replace NeRF with 3D Gaussian Splatting to accelerate optimization while maintaining competitive visual quality. 
Other studies \cite{liu2026hifi3d,zhou2026consdreamer,qin2026distilling} further improve multi-view consistency by adapting pretrained diffusion models for view synthesis. 
While these models can generate visually appealing results, their iterative optimization is computationally expensive and may suffer from multi-view inconsistency and geometric artifacts due to the lack of explicit 3D supervision. 
In contrast, feedforward-based models directly learn from large-scale 3D benchmarks \cite{deitke2023objaversexl,deitke2023objaverse,su2024gt23d} to generate 3D assets. Representative approaches include Point-E \cite{nichol2022pointe}, Shap-E \cite{jun2023shape}, and other 3D generators \cite{chen2025lara,wu2026animateanymesh++,tang2024lgm}.
More recently, flow matching \cite{lipman2022flow} has emerged as an effective formulation for improving sampling efficiency in feedforward-based models \cite{miao2026dreamer,sun2026meshflow,li2025triposg,fei2026home3d}. TRELLIS \cite{xiang2025trellis} combines structured 3D latent representations with flow matching to achieve efficient sampling and high-quality geometry and appearance. Hunyuan3D-2.1 \cite{hunyuan3d21} adopts a dual-branch framework for shape and texture generation, while Meshy T2 \cite{xu2026meshyt2} combines voxel-level and mesh-level flow models in a coarse-to-fine framework, enabling high-fidelity generation with substantially reduced inference time. Overall, flow-matching-based feedforward models achieve a favorable trade-off between generation efficiency and fidelity, providing a suitable foundation for subsequent RL optimization.

\subsection{RL Training Strategy}

\textbf{RL Algorithms for Diffusion Models.} 
Inspired by the success of RL in LLMs \cite{shao2024deepseekmath,li2026enhancing,liu2026mracl}, recent studies have extended RL-based post-training to diffusion and flow-based generative models \cite{liu2026advancesgrpogenerationmodels}. Early approaches optimize the diffusion score function with policy-gradient objectives \cite{black2024training,zhang2026aligning}, enabling generated samples to better align with human preferences. Subsequently, \cite{wallace2024diffusion} introduces offline Direct Preference Optimization for the first time. This allows diffusion models to directly learn from human feedback and validates its effectiveness on large-scale models. Some works \cite{liu2026improving,luo2026reinforcing,peng2026self} have adopted online methods, continuously adjusting sampling trajectories through step-aware preference models during training to achieve improved performance. More recently, the success of Group Relative Policy Optimization (GRPO) \cite{shao2024deepseekmath} has motivated group-based relative optimization for diffusion models \cite{liu2026flowgrpo,ding2026treegrpo,rojas2026improving}. Representative methods such as Flow-GRPO \cite{liu2026flowgrpo} and DanceGRPO \cite{xue2025dancegrpo} extend GRPO to flow-matching models by introducing stochastic sampling trajectories, enabling group-wise advantage estimation without an explicit critic model. Although effective, these methods still involve trajectory-level sampling and optimization, which can incur substantial computational and memory overhead for diffusion and flow models. To further improve training efficiency, DiffusionNFT \cite{zheng2026diffusionnft} replaces trajectory-level optimization over the discretized reverse process with reward-guided fine-tuning through the forward process, avoiding the storage of full sampling trajectories. AWM \cite{xue2025awm} further simplifies optimization by directly incorporating advantage weights into the native score/flow-matching objective. Both methods improve the computational efficiency of diffusion RL by reducing trajectory-level computation and memory overhead.

\textbf{Multi-Dimensional Optimization for RL.} 
Multi-dimensional rewards are commonly used to capture complementary aspects of generation quality. However, averaging these reward signals into a single objective can induce dimension dominance, biasing policy optimization toward a subset of quality dimensions at the expense of others \cite{li2026enhancing}. FocalReward \cite{huang2026focal} addresses this issue by characterizing reward polarization across evaluation dimensions and dynamically reweighting rewards according to their saturation levels. Multi-GRPO \cite{lyu2025multi} further alleviates dimension dominance and interference by independently normalizing per-reward advantages before aggregation.

Beyond reward imbalance, multi-dimensional optimization also suffers from inter-dimension conflicts, as different objectives may induce inconsistent or even opposing gradient directions \cite{liu2021conflict,rame2023rewarded}. To resolve such conflicts, prior studies have explored gradient-level optimization strategies, including identifying common descent directions across objectives \cite{desideri2012multiple,sener2018multi}, projecting conflicting gradient components \cite{yu2020gradient}, and balancing per-objective improvements \cite{liu2021conflict}. These techniques have also been introduced to diffusion models. Parrot \cite{lee2024parrot} formulates image generation as a multi-objective RL problem and employs Pareto-optimal selection to balance visual fidelity, textual alignment, and human preferences. T2I-R1 \cite{jiang2026t2i} integrates semantic- and token-level rewards to jointly guide high-level planning and fine-grained image generation. More recently, AR3D-R1 \cite{tang2026we} introduces Hi-GRPO to progressively optimize global and local 3D properties using their corresponding reward signals. MARBLE \cite{zhao2026marble} separately computes advantages for individual rewards and coordinates their gradients to mitigate conflicts in multi-dimensional diffusion RL.

Despite these advances, existing methods mainly mitigate multi-dimensional interference through reward reweighting or gradient-level balancing, while still optimizing all dimensions simultaneously within each policy update. They do not explicitly model the dependencies among dimensions, which may lead to persistent conflicts and limit the improvement of other dimensions. This motivates an interference-aware sequential reward scheduling strategy that determines the optimization order and adaptively controls transitions between dimension-wise reward objectives.

\section{Methodology}\label{sec:methodology}

\noindent Motivated by the aforementioned limitations of existing multi-dimensional RL methods, we propose a unified framework consisting of two complementary components: \textbf{AdaSelect} and \textbf{OuroReward}. Given $K$ quality dimensions such as \textit{Alignment} and \textit{Geometry}, we associate each dimension with a corresponding reward, denoted as $\{R^{(k)}\}_{k=1}^{K}$. AdaSelect periodically updates the selected prompts during training to keep them aligned with the model’s evolving capability and provide informative learning signals, thereby improving training stability and effectiveness. To address inter-dimension interference, OuroReward estimates dependencies among dimensions and derives a cyclic optimization path that minimizes cumulative interference. Then, multiple dimension-wise rewards are optimized along the derived order, where the reward objective is adaptively switched from one to the next based on the current dimension's optimization headroom. In this way, the proposed framework improves both the generation quality and training stability of multi-dimensional RL optimization. In the following, Sec.~\ref{sec:adaselect} introduces the proposed AdaSelect, Sec.~\ref{sec:OuroReward} presents OuroReward, and Sec.~\ref{sec:training_policy} describes the training policy used for RL optimization.

\begin{figure*}[!t]
    \centering
    \includegraphics[width=0.95\linewidth]{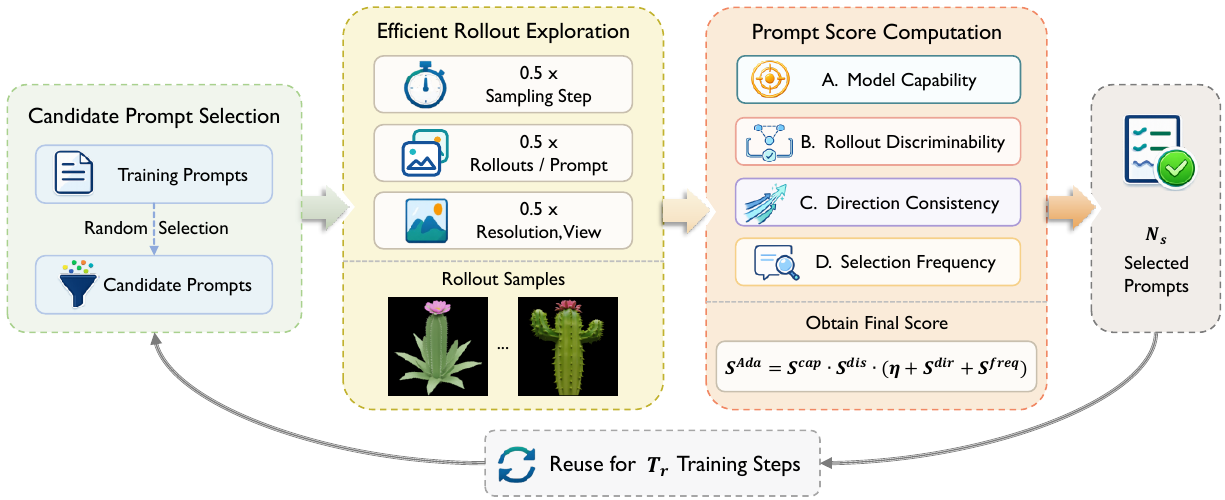}
    \caption{The framework of the proposed AdaSelect strategy.}
    \label{fig:framework_ada}
    \vspace{-0.4cm}
\end{figure*}

\subsection{AdaSelect: Adaptive Sample Selection}\label{sec:adaselect}

The overall framework of the proposed AdaSelect is illustrated in Fig.~\ref{fig:framework_ada}, which consists of three components: \textit{Candidate Prompt Selection}, \textit{Efficient Rollout Exploration}, and \textit{Prompt Score Computation}. The \textit{Candidate Prompt Selection} module first constructs a candidate pool from the training set every few training steps. \textit{Efficient Rollout Exploration} then performs lightweight 3D generation for each candidate prompt, after which the \textit{Prompt Score Computation} module evaluates each prompt based on its generated 3D samples and the proposed scoring criteria. Finally, the candidate prompts are ranked by their scores, and those most suitable for the current training stage are selected for subsequent RL optimization.

\subsubsection{Candidate Prompt Selection} 

Given a training set $\mathcal{P}_{train}$ containing $N_{train}$ prompts, AdaSelect first constructs a compact candidate pool $\mathcal{P}_{cand}$ to limit the scope of subsequent prompt selection. We set the candidate pool size to $N_{c}=N_{train}/10$ and uniformly sample prompts without replacement from $\mathcal{P}_{train}$:
\begin{equation}
\mathcal{P}_{cand}\sim\mathrm{RandomSample}\left(\mathcal{P}_{train},N_{c}\right).
\end{equation}

\subsubsection{Efficient Rollout Exploration} 

After constructing the candidate pool, AdaSelect performs lightweight rollout exploration for each prompt to generate 3D samples for subsequent prompt scoring. To reduce the evaluation cost, we adopt a lightweight inference configuration, where the number of sampling steps $T_{c}$, group size $G_{c}$, rendered image resolution $L_{c}$, and number of viewpoints $V_{c}$ are each set to 50\% of their corresponding training configurations (denoted by $T_{train}$, $G_{train}$, $L_{train}$ and $V_{train}$):
\begin{align}
    T_{c}&=0.5T_{train}, \  G_{c}=0.5G_{train},\\
    L_{c}&=0.5L_{train}, \ V_{c}=0.5V_{train}.
\end{align}
This lightweight configuration reduces rollout-generation cost while preserving the key geometric and appearance characteristics of the generated 3D content. Consequently, the resulting samples provide sufficient information to assess the relative suitability of candidate prompts for subsequent selection.

\subsubsection{Prompt Score Computation}\label{sec:prompt_score_computation}

Given the 3D samples generated through efficient rollout exploration, we apply reward models to evaluate each candidate prompt across multiple dimensions. The reward vector of the $m$-th rollout for the $n$-th prompt at training step $t$ is defined as:
\begin{equation}
\mathbf{R}_{n,m,t}=\left[R_{n,m,t}^{(1)},R_{n,m,t}^{(2)},\ldots,R_{n,m,t}^{(K)}\right],
\end{equation}
where $K$ denotes the number of quality dimensions and ${R}_{n,m,t}^{(k)}$ denotes the reward associated with the $k$-th dimension.

\begin{figure}[!t]
    \centering
    \includegraphics[width=0.98\linewidth]{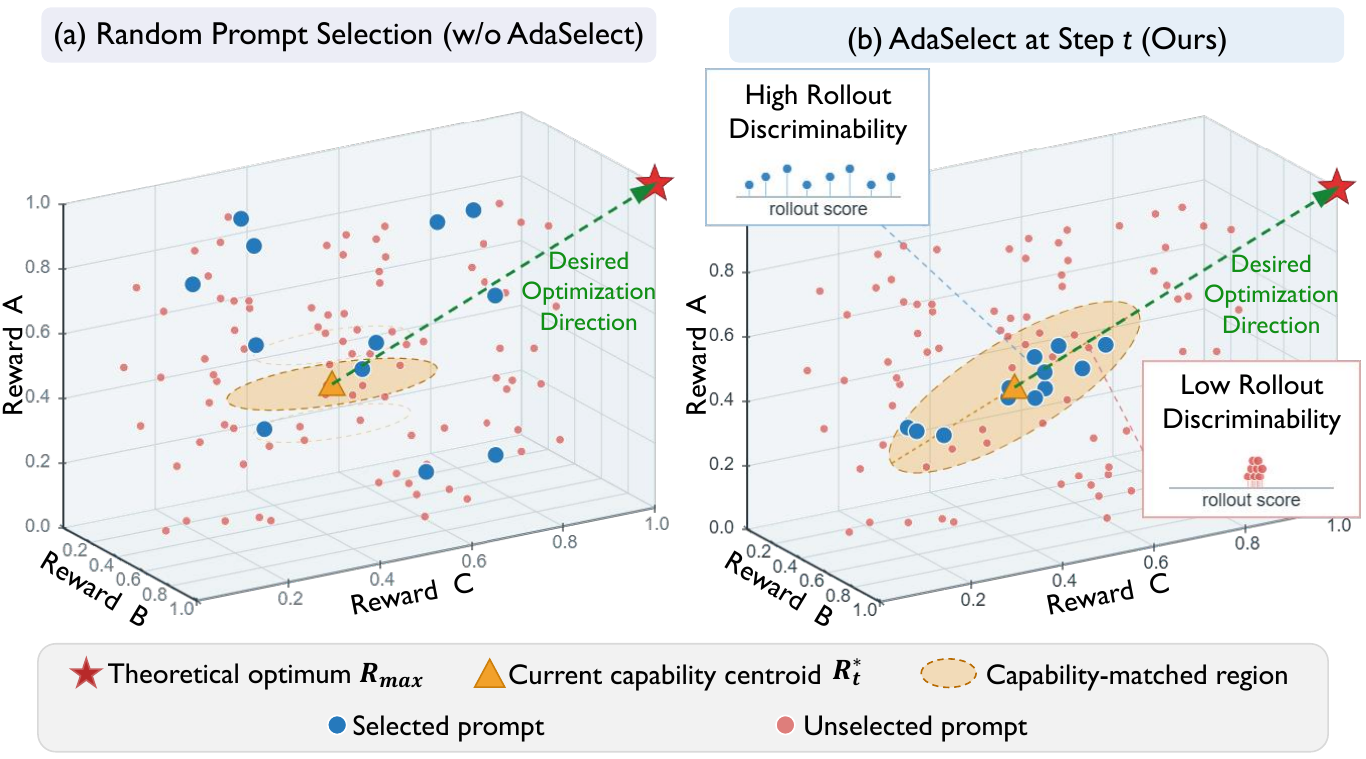}
    \caption{Comparison of Random and AdaSelect for Prompt Selection at Step $t$.}
    \label{fig:prompt_selection_process}
    \vspace{-0.4cm}
\end{figure}

\textbf{Model Capability.} Inspired by MRACL \cite{liu2026mracl}, selecting prompts that match the model's current capability is important for stable and effective optimization. For each candidate prompt, we first aggregate the reward vectors over its $G_{c}$ rollouts to obtain a prompt-level reward representation:
\begin{equation}
\mathbf{R}_{n,t}=\frac{1}{G_{c}}\sum_{m=1}^{G_{c}}\mathbf{R}_{n,m,t}.
\end{equation}
Based on \cite{liu2026mracl}, we characterize the model's current capability using the centroid of $\mathbf{R}_{n,t}$ over the candidate set:
\begin{equation}
\mathbf{R}_{t}^{*}=\frac{1}{N_{c}}\sum_{n=1}^{N_{c}}\mathbf{R}_{n,t}.
\end{equation}
We then measure the compatibility of the $n$-th prompt with the current model capability by its Euclidean distance to the centroid and further transform the resulting distance into a normalized capability score using a Gaussian kernel:
\begin{equation}
d^{cap}_{n,t}=\left \| \mathbf{R}_{n,t} - \mathbf{R}_{t}^{*}\right \| _2,
\end{equation}
\begin{equation}
S_{n,t}^{cap}=\exp\left(-\frac{(d^{cap}_{n,t})^2}{2\sigma_{d,t}^2}\right),
\end{equation}
where $\sigma_{d,t}^2$ denotes the variance of the capability distances across all candidate prompts at training step $t$. A higher $S_{n,t}^{cap}$ indicates that the corresponding prompt is more closely aligned with the model's current capability and is therefore more suitable for the current curriculum stage.

\textbf{Rollout Discriminability.} Following \cite{li2026enhancing, liu2026flowgrpo}, effective RL optimization relies on sufficient reward variation among group rollouts for the same prompt. When the rollouts generated from a prompt receive nearly identical rewards, they provide weak comparative signals for policy optimization. Therefore, we select prompts based on their rollout variation. Specifically, for the $n$-th candidate prompt, we define its discriminability as the average reward variation across the $K$ dimensions:
\begin{equation}
D_{n,t}=\frac{1}{K}\sum_{k=1}^{K}\operatorname{Std}_{m=1,\ldots,G_{c}}
\left(R_{n,m,t}^{(k)}\right).
\end{equation}
We then normalize the resulting value over the candidate set to obtain the rollout discriminability score:
\begin{equation}
S_{n,t}^{dis}=\frac{D_{n,t}-D_{n,t}^{\text{min}}}{D_{n,t}^\text{max}-D_{n,t}^\text{min}}.
\end{equation}
A higher $S_{n,t}^{dis}$ indicates that the corresponding prompt produces more diverse group rollouts, providing sufficiently discriminative reward signals for relative policy optimization.

\textbf{Direction Consistency.} For multi-dimensional optimization, prompts with comparable aggregate rewards may exhibit distinct dimension-wise reward distributions. To account for such differences, we further introduce direction consistency to evaluate whether the optimization direction of each prompt is consistent with the desired multi-dimensional direction. We first define the theoretical maximum reward vector across the $K$ dimensions as:
\begin{equation}
\mathbf{R}_{\max}=\left[R_{\max}^{(1)},\ldots,R_{\max}^{(K)}\right],
\end{equation}
where $R_{\max}^{(k)}$ denotes the theoretical upper bound of the $k$-th reward. Accordingly, the optimization direction of the $n$-th prompt is then obtained by:
\begin{equation}
\mathbf{O}_{n,t}=\mathbf{R}_{\max}-\mathbf{R}_{n,t}.
\end{equation}
Using the current capability centroid $\mathbf{R}_{t}^{*}$ as the reference, we derive the desired multi-dimensional direction as:
\begin{equation}
\mathbf{O}_{t}^{*}=\mathbf{R}_{\max}-\mathbf{R}_{t}^{*}.
\end{equation}
We then quantify the alignment between $\mathbf{O}_{n,t}$ and $\mathbf{O}_{t}^{*}$, and further normalize the resulting value over the candidate set to obtain the direction consistency score:
\begin{equation}
c_{n,t}^{dir}=\frac{\mathbf{O}_{n,t}^{\top}\mathbf{O}_{t}^{*}}{
\left \| \mathbf{O}_{n,t} \right \| \left \| \mathbf{O}_{t}^{*} \right \|},
\end{equation}
\begin{equation}
S_{n,t}^{dir}=\frac{c_{n,t}^{dir}-c_{\min,t}^{dir}}{c_{\max,t}^{dir}-c_{\min,t}^{dir}}.
\end{equation}
A higher $S_{n,t}^{dir}$ indicates stronger alignment with the desired multi-dimensional direction, suggesting that the corresponding prompt is more suitable for subsequent optimization.

\textbf{Selection Frequency.} Repeatedly selecting the same prompts may limit training diversity and overemphasize a small subset of the training data. To encourage broader prompt coverage, we introduce a frequency-aware selection score:
\begin{equation} 
S_{n,t}^{freq}=\left(1+\frac{f_{n,t}}{f_s}\right)^{-\alpha_s},
\end{equation}
where $f_{n,t}$ denotes the selection times of the $n$-th prompt before step $t$, $f_s$ is a scaling factor, and $\alpha_s$ controls the decay rate. A smaller $f_{n,t}$ yields a larger $S_{n,t}^{freq}$, making the corresponding prompt more likely to be selected.

These four components evaluate candidate prompts from complementary perspectives. Model capability and rollout discriminability serve as the primary criteria, indicating whether a prompt provides appropriate and reliable learning signal for the current training stage. Direction consistency and selection frequency provide auxiliary information to further distinguish prompts that satisfy these primary conditions. Accordingly, the overall prompt score is defined as:
\begin{equation}
S_{n,t}^{Ada}
=S_{n,t}^{cap} \cdot S_{n,t}^{dis}\cdot \left(\eta +S_{n,t}^{dir}+S_{n,t}^{freq}\right).
\end{equation}
The multiplicative formulation emphasizes capability and discriminability as the dominant factors, while $\eta$ controls the relative influence of the auxiliary components.

At training step $t$, the $N_s$ prompts with the highest $S_{n,t}^{\mathrm{Ada}}$ are selected as the active training subset. Fig.~\ref{fig:prompt_selection_process} illustrates the prompt selection process. This subset is used for the next $T_r$ RL steps, during which prompts are sampled from it for rollout generation and policy optimization. Afterward, the candidate pool is refreshed and rescored to construct a new active subset. This periodic update enables prompt selection to remain aligned with the model's evolving capability.

\begin{figure*}[!t]
    \centering
    \includegraphics[width=0.98\linewidth]{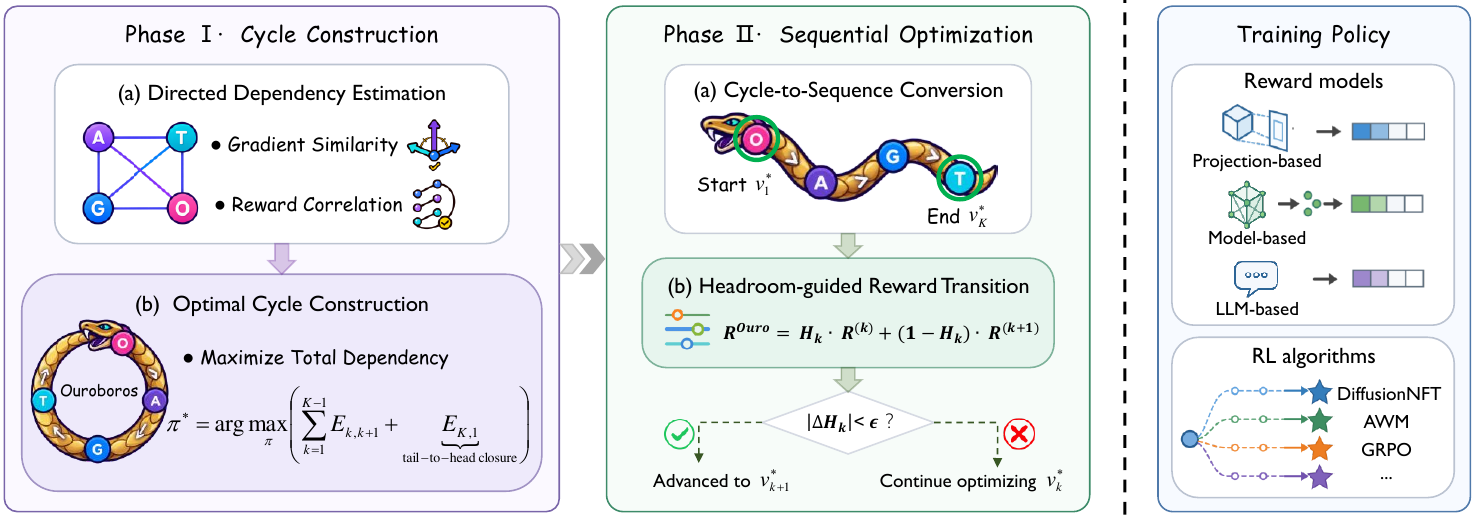}
    \caption{The framework of the proposed OuroReward strategy and the RL training policy.}
    \label{fig:framework_OuroReward}
    \vspace{-0.4cm}
\end{figure*}

\subsection{OuroReward: Interference-Aware Sequential Reward Scheduling}\label{sec:OuroReward}

To account for the heterogeneous dependencies among dimension pairs and mitigate the interference induced by simultaneous optimization, we propose \textbf{OuroReward}, a sequential reward scheduling strategy for T23D RL. OuroReward consists of two components: \textit{Multi-Dimensional Cycle Construction} and \textit{Dependency-Guided Sequential Optimization}. In OuroReward, each dimension is characterized by its corresponding reward, and the optimization process is organized around these dimension-wise reward objectives. As illustrated in Fig.~\ref{fig:framework_OuroReward}, \textit{Multi-Dimensional Cycle Construction} first estimates directed dependencies among dimension pairs using gradient cosine similarity and reward-change correlation, and then constructs an optimal dependency path that maximizes overall inter-dimension compatibility. The cyclic formulation explicitly incorporates the \textit{tail-to-head} dependency into path evaluation, enabling the cycle to capture global compatibility beyond local adjacent transitions. Given the constructed cycle, \textit{Dependency-Guided Sequential Optimization} selects the most compatible dimension as the starting target. The cycle is then unrolled from this target into a one-pass training sequence. During training, the reward objective is adaptively adjusted according to the optimization headroom. In this way, OuroReward replaces simultaneous optimization with interference-aware sequential scheduling, thereby mitigating cumulative inter-dimension interference and stabilizing T23D RL training.

\subsubsection{Multi-Dimensional Cycle Construction}

To determine an interference-aware optimization order, we first construct a directed dependency graph over the $K$ dimensions, where each node $v_k$ represents a dimension and each directed edge quantifies the influence of optimizing one on another. We then conduct $K$ independent lightweight warm-up runs, all initialized from the same policy parameters. In the $k$-th run, the policy is optimized exclusively using the $k$-th dimension-wise reward for $T_w$ training steps. At each step $t$, we compute the policy loss $\mathcal{L}_{k,t}^{\mathrm{policy}}$ and record the reward score $R_{n,t}^{(k)}$ for all $K$ dimensions on a validation set with $N_{val}$ prompt. Using the policy loss and reward trajectory from the $k$-th run, we compute the dependency score $E_{k,l}$ to characterize the directed influence of optimizing the $k$-th reward on the $l$-th reward. Conversely, $E_{l,k}$ is computed from the $l$-th run and represents the influence of optimizing the $l$-th reward on the $k$-th reward. Therefore, $E_{k,l}$ and $E_{l,k}$ capture the two directions of dependency between the two dimensions. Finally, we identify a cyclic path with the highest cumulative dependency score, which corresponds to lower cumulative interference during optimization.

\textbf{Directed Dependency Estimation.} The dependency score $E_{k,l}$ is computed from two complementary measures: \textit{Gradient Similarity} and \textit{Reward Correlation}. Gradient Similarity evaluates the alignment between the optimization directions of the $k$-th and $l$-th dimension-wise rewards. Specifically, given the policy loss $\mathcal{L}_{k,t}^{\mathrm{policy}}$ associated with the $k$-th reward, we can compute its gradient as:
\begin{equation}
\mathbf{g}_{k,t}=\nabla_{\boldsymbol{\theta}}\mathcal{L}_{k,t}^{\mathrm{policy}}.
\end{equation}
For gradients $\mathbf{g}_{k,t}$ and $\mathbf{g}_{l,t}$ from the $k$-th run, we compute their cosine similarity and accumulate the gradient similarity over all $T_w$ training steps to reduce sensitivity to noise in individual updates:
\begin{equation}
E_{k,l}^{sim}=\sum_{t=1}^{T_{w}}\frac{\mathbf{g}_{k,t}^{\top}\cdot \mathbf{g}_{l,t}}{\left \| \mathbf{g}_{k,t} \right \| \left \| \mathbf{g}_{l,t} \right \|}.
\end{equation}
A larger $E_{k,l}^{sim}$ indicates stronger policy-update direction similarity, whereas a negative value indicates potential gradient conflict.

Reward Correlation evaluates the consistency of reward changes across consecutive steps on the validation set $N_{val}$, which is quantified by the PLCC between the changes of two reward objectives. For each transition from step $t-1$ to step $t$ during the $k$-th run, we construct the sample-wise reward-change vector of $k$-th reward as:
\begin{equation}
\Delta\mathbf{R}_{t}^{(k)}
=\{R_{n,t}^{(k)}-R_{n,t-1}^{(k)}\}_{n=1}^{N_{val}}.
\end{equation}
Similarly, we compute the step-wise reward correlation between the $k$-th and $l$-th rewards from the $k$-th run and accumulate the step-wise correlations over all consecutive steps:
\begin{equation}
E_{k,l}^{corr}=\sum_{t=2}^{T_w}\mathrm{PLCC}\left(\Delta \mathbf{R}_{t}^{(k)},\Delta \mathbf{R}_{t}^{(l)}\right).
\end{equation}
A larger $E_{k,l}^{corr}$ indicates stronger correlation between the two rewards. The final dependency score is obtained by averaging the two measures:
\begin{equation}
E_{k,l}= \frac{E_{k,l}^{sim}+E_{k,l}^{corr}}{2}.
\end{equation}

\textbf{Optimal Cycle Construction.} Given the dependency score $E_{k,l}$ for each directed edge, we evaluate a candidate cyclic path $\pi=(v_{1},v_{2},\ldots,v_{K})$ by accumulating the dependency scores between adjacent dimensions, including the \textit{tail-to-head} edge, where $v_k$ denotes the $k$-th dimension along the path:
\begin{equation}
S\left ( \pi \right )=\sum_{k=1}^{K-1} E_{k,k+1}+\underset{\text{tail-to-head closure}}{\underbrace{E_{K,1}}}.
\end{equation}
A larger $S(\pi)$ indicates stronger overall compatibility and lower cumulative interference. The optimal cyclic path is therefore selected as:
\begin{equation}
\pi^{*}=\arg\max_{\pi\in\Pi } S\left ( \pi \right ),
\end{equation}
where $\Pi$ denotes the set of all candidate cyclic paths. Unlike greedy transition selection, this formulation evaluates the complete cyclic structure and avoids locally optimal but globally suboptimal orders.

\subsubsection{Dependency-Guided Sequential Optimization} 

After constructing the optimal dependency cycle $\pi^{*}$, OuroReward converts it into a concrete one-pass training sequence and performs headroom-guided reward transitions along the sequence.

\textbf{Cycle-to-Sequence Conversion.} The dependency cycle determines the relative ordering among rewards, but its rotational invariance leaves the starting point undetermined. Starting from a highly conflicting dimension-wise reward may propagate interference through subsequent updates and destabilize training. Therefore, we select the reward with the highest aggregate dependency score to all other dimensions as the starting point:
\begin{equation}
v_{start}^{*}=\arg\max_{v_k\in\pi^{*}}\sum_{l\ne k} E_{k,l}.
\end{equation}
OuroReward then unfolds $\pi^{*}$ to begin with $v_{start}^{*}$, yielding a one-pass sequence for dimension-wise optimization.

\textbf{Headroom-guided Reward Transition.} 
Given the resulting training sequence, OuroReward prevents over-optimization of any single reward by adaptively transitioning the reward objective according to the optimization headroom of the current step. During training, we generate $G_{\mathrm{train}}$ rollouts for each prompt. The reward vector of the $m$-th rollout is represented as:
\begin{equation}
\mathbf{R_m}=\left [ R_m^{(1)},  R_m^{(2)},\dots R_m^{(K)}\right ], \ m\in\{1,\ldots,G_{train}\}.    
\end{equation}
To mitigate scale discrepancies across dimensions, we first normalize each reward:
\begin{equation}
\widetilde{R}_m^{(k)}
=\frac{R_m^{(k)}-R_{\text{min}}^{(k)}}{R_{\text{max}}^{(k)}-R_{\text{min}}^{(k)}},
\end{equation}
where $R_{\text{min}}^{(k)}$ and $R_{\text{max}}^{(k)}$ denote the lower and upper bounds of the $k$-th dimension-wise reward, respectively. Following \cite{huang2026focal}, we first average the rewards and then convert the resulting rewards into Gibbs-normalized weights:
\begin{equation}
R_m^{base}=\frac{1}{K}\sum_{k=1}^{K} \widetilde{R}_m^{(k)}, \quad \omega_m
=
\frac{\exp\left(R_m^{base}/\tau\right)}
{\sum\exp\left(R_m^{base}/\tau\right)},
\end{equation}
where $\tau$ denotes the temperature parameter. The optimization headroom of the current $k$-th dimension-wise reward is then computed as:
\begin{equation}
H_k=1-
\sum_{m=1}^{G_{\mathrm{train}}}
\omega_m \cdot
\widetilde{R}_m^{(k)}.
\end{equation}
A larger $H_k$ indicates greater remaining optimization potential. More details of the headroom estimation are provided in \cite{huang2026focal}.

To enable a smooth transition between adjacent reward objectives, OuroReward constructs the training reward by adaptively weighting the current dimension-wise reward and the next reward along the optimal path:
\begin{equation}
R^{\mathrm{Ouro}}
= H_k R^{(k)} + (1-H_k) R^{(k^+)},
\quad
k^+ =
\begin{cases}
k+1, & k < K,\\
1, & k = K.
\end{cases}
\end{equation}
where $R^{(k)}$ and $R^{(k^+)}$ denote the rewards associated with $v_k^{*}$ and $v_{k^+}^{*}$, respectively. The adaptive coefficient $H_k$ controls the transition from the current reward objective to the next one. When $|\Delta H_k|<\epsilon$ across consecutive training steps, the current reward is considered stable, and training proceeds to the next reward $R^{(k^+)}$. After the terminal reward objective $R^{(K)}$ is reached, the schedule follows the \textit{tail-to-head} transition and returns to the starting reward $R^{(1)}$, which is then optimized until the end of training. This one-pass scheduling strategy enables smooth reward transitions while reducing persistent interference caused by simultaneous multi-dimensional optimization.

\begin{figure*}[!t]
    \centering
    \begin{subfigure}{0.26\linewidth}
        \centering
        \includegraphics[width=\linewidth]{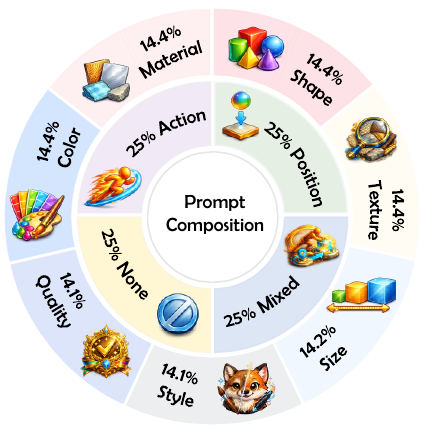}
        \caption{}
        \label{fig:prompt_composition}
    \end{subfigure}
    \hfill
    \begin{subfigure}{0.34\linewidth}
        \centering
        \includegraphics[width=\linewidth]{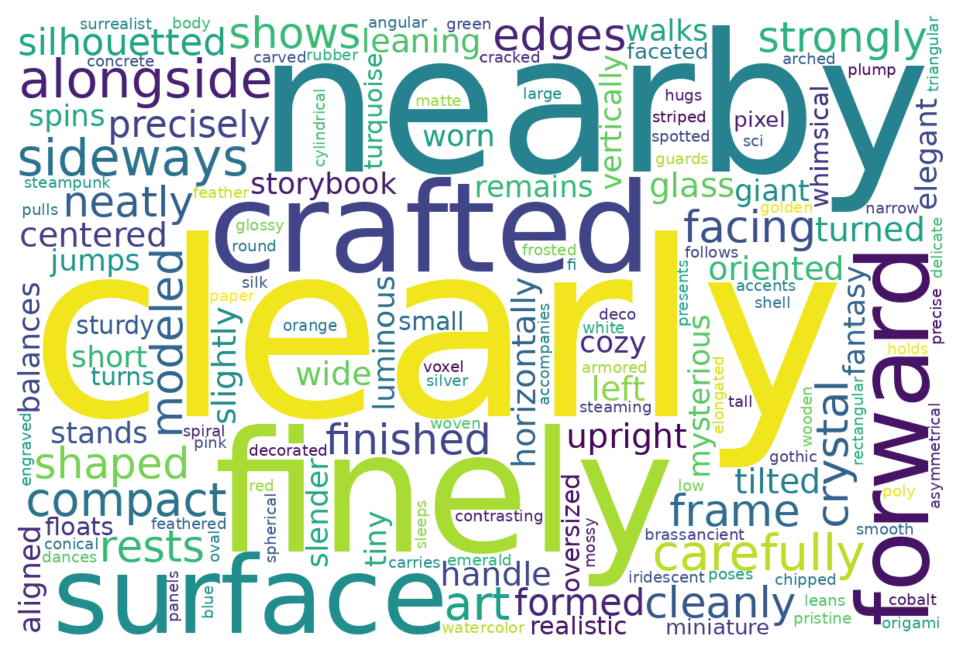}
        \caption{}
        \label{fig:word_cloud}
    \end{subfigure}
    \hfill
    \begin{subfigure}{0.36\linewidth}
        \centering
        \includegraphics[width=\linewidth]{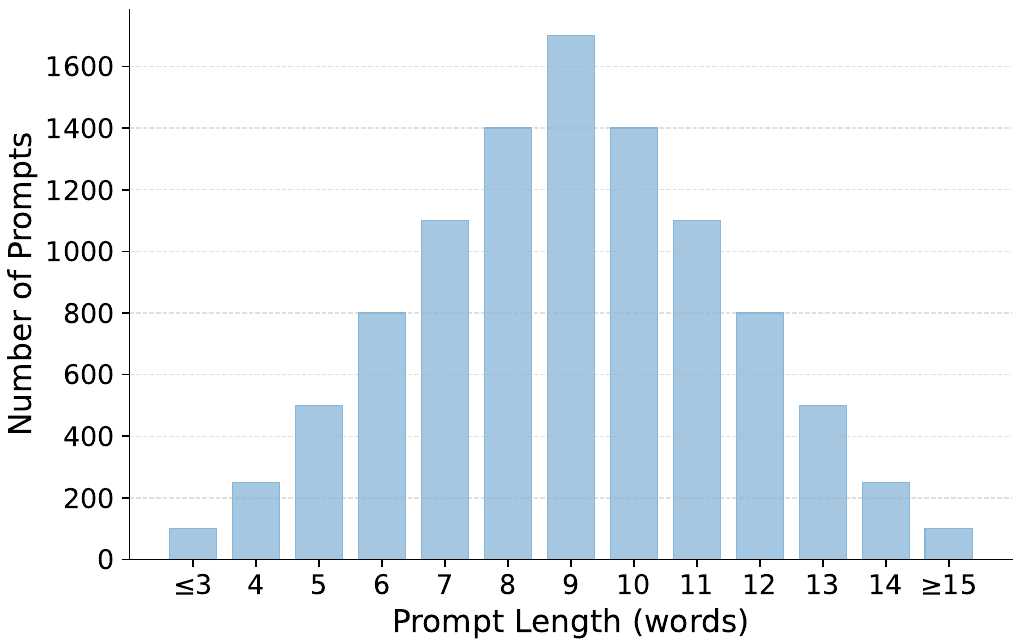}
        \caption{}
        \label{fig:prompt_length}
    \end{subfigure}
    \caption{Statistical analysis of the constructed large-scale prompt dataset. (a) Prompt composition, where the inner and outer circles represent interaction types and attribute categories, respectively. (b) Word cloud visualization of the prompt corpus. (c) Distribution of prompt lengths.}
    \label{fig:prompt_statistical_analysis}
    \vspace{-0.4cm}
\end{figure*}

\subsection{Training Policy}\label{sec:training_policy}

The proposed framework is compatible with diverse reward models and RL algorithms, which provide reward supervision and policy optimization, respectively. As illustrated in Fig.~\ref{fig:framework_OuroReward}, our framework supports diverse reward models, including projection-based, model-based, and LLM-based evaluators. Detailed descriptions of these reward models are provided in the Appendix. Given a generated 3D sample, each reward model evaluates a specific quality dimension and produces a scalar reward. The resulting multi-dimensional rewards are used throughout the framework. AdaSelect leverages these rewards to estimate prompt informativeness and select training prompts that are suitable for the current model state, while OuroReward uses them to estimate inter-dimension dependencies and guide interference-aware sequential optimization.

During policy optimization, the reward objective depends on the current scheduled dimension and updates the generative model via an online RL algorithm. OuroReward only controls the reward scheduling process and is independent of the specific policy optimizer. Therefore, it can be readily integrated with different RL algorithms, including DiffusionNFT \cite{zheng2026diffusionnft}, GRPO \cite{liu2026flowgrpo}, and AWM \cite{xue2025awm}, demonstrating its flexibility across different RL training paradigms.

\section{Experiments}\label{sec:experiments}

\subsection{Datasets and Evaluation Metrics} \label{sec:datasets_evaluation}

Existing T23D benchmarks \cite{zhou20253dgcqa,he2023t3bench,wu2024gpt4v,zhang2025hyperscore,cui2026rank2score,su2024gt23d} typically contain relatively limited prompt sets, which are insufficient for RL optimization. To address this limitation, we employ GPT-5.6 \cite{openaigpt} to construct a large-scale prompt dataset for training, validation, and test. Specifically, we collect prompts from existing benchmarks, including 3DGCQA \cite{zhou20253dgcqa}, T3Bench \cite{he2023t3bench}, GPTEval \cite{wu2024gpt4v}, MATE-3D \cite{zhang2025hyperscore}, T23D-CompBench \cite{cui2026rank2score}, and T23DAQA \cite{su2024gt23d}. These prompts are provided to GPT-5.6 with instructions to remove duplicates, correct incomplete or invalid prompts, and expand the dataset following a compositional schema consisting of subjects, attributes, materials, actions, relationships, scenes, and styles. During this process, duplicate prompts and semantically implausible combinations are filtered to improve the diversity and validity of the generated dataset. The detailed instruction template is illustrated in the Appendix. The generated dataset contains 10,000 prompts, which are randomly split into $N_{train}=8,000$ training prompts, $N_{val}=1,000$ validation prompts and $N_{test}=1,000$ test prompts. As shown in Fig.~\ref{fig:prompt_statistical_analysis}, the constructed dataset covers diverse interaction types, attribute categories, and compositional complexities, providing a comprehensive prompt space for RL optimization.

Following prior studies \cite{xiang2025trellis,zheng2026diffusionnft,lai2026lattice}, we evaluate T23D generative models on the constructed test set. For each prompt, we generate $G_{\mathrm{test}}=8$ rollouts and average their scores to obtain the prompt-level result. For objective assessment, we employ a diverse set of evaluators, including projection-based (i.e., CLIPScore \cite{liu2025clip}, ImageReward \cite{xu2024imagereward}, HyperScore \cite{zhang2025hyperscore}, and Rank2Score \cite{cui2026rank2score}), model-based (i.e., Uni3D \cite{zhou2024uni3d}), and LLM-based methods (i.e., Qwen3-VL-8B-Instruct \cite{bai2025qwen2}). The final score of each metric is computed by averaging the results over all $N_{\mathrm{test}}$ test prompts. In addition to objective evaluation, we conduct a user study to assess human preference. We randomly sample 300 prompts from the test set and generate the corresponding 3D assets using each compared model. For each prompt, 10 independent participants rank the generated assets, and then we aggregate these rankings to compute each model's win rate.

\begin{table*}[t]
\centering
\caption{Performance comparison of our method with representative baselines across proxy rewards and unseen evaluators. Our method comprises two components: AdaSelect and OuroReward. The proxy reward is marked in orange. The best result in each column is shown in \textbf{bold}, while the second is \underline{underlined}.}
\setlength{\abovecaptionskip}{-2pt}
\label{tab:overall_performance1}
\begin{tabular}{l|cc|ccccc}
\toprule
\multirow{2}{*}{Model} & \multicolumn{2}{c|}{Trained Tasks}& \multicolumn{5}{c}{Unseen Tasks} \\ \cmidrule{2-8}
& HyperScore & Rank2Score  & CLIPScore& ImageReward & Uni3D & \makecell{Qwen3-VL-8B-\\Instruct} & \cellcolor{lightgreen}Win Rate
\\
\midrule
\textcolor{softblack}{TRELLIS} & \textcolor{softblack}{3.466}&\textcolor{softblack}{3.713}&\textcolor{softblack}{0.208}&\textcolor{softblack}{0.362}&\textcolor{softblack}{0.254}&\textcolor{softblack}{6.483}& \cellcolor{lightgreen}\textcolor{softblack}{-} \\ \cmidrule{1-8}
\multicolumn{8}{c}{\textit{HyperScore Task}} \\
+ GRPO & \cellcolor{orange!20}3.782&3.696 &0.205 &0.364 &0.252 &6.479&\cellcolor{lightgreen}0.007  \\ 
+ DiffusionNFT & \cellcolor{orange!20}4.050&3.798 &0.213 &0.467  &\underline{0.261} &6.671 &\cellcolor{lightgreen}0.033 \\ 
\rowcolor{gray!15} + GRPO + \textbf{Ours} & \underline{\cellcolor{orange!20}4.278} & \underline{3.861}&\underline{0.219} &\underline{0.598} &0.260 &\underline{6.710} &\cellcolor{lightgreen}\underline{0.337}\\ 
\rowcolor{gray!15}+ DiffusionNFT + \textbf{Ours}& \textbf{\cellcolor{orange!20}4.483} &\textbf{3.937} &\textbf{0.228} &\textbf{0.662} &\textbf{0.266} &\textbf{6.835} &\cellcolor{lightgreen}\textbf{0.623}\\ \cmidrule{1-8}
\multicolumn{8}{c}{\textit{Rank2Score Task}} \\
+ GRPO &3.463& \cellcolor{orange!20}4.108&0.206 &0.367 &0.250 & 6.636& \cellcolor{lightgreen}0.010\\ 
+ DiffusionNFT & 3.575&\cellcolor{orange!20}4.253 &0.216 &0.483 & 0.253 &6.781& \cellcolor{lightgreen}0.020\\ 
\rowcolor{gray!15} + GRPO + \textbf{Ours} &\underline{3.613} &\underline{\cellcolor{orange!20}4.392} &\underline{0.222} & \underline{0.601}& \underline{0.262}&\underline{6.860} &\cellcolor{lightgreen}\underline{0.317}\\
\rowcolor{gray!15}+ DiffusionNFT + \textbf{Ours}&\textbf{3.654} & \textbf{\cellcolor{orange!20}4.525}& \textbf{0.233}&\textbf{0.678} &\textbf{0.267} &\textbf{6.982} &\cellcolor{lightgreen}\textbf{0.653}\\ \midrule
\textcolor{softblack}{Hunyuan3D-2.1} & \textcolor{softblack}{3.625}&\textcolor{softblack}{4.151}&\textcolor{softblack}{0.232}&\textcolor{softblack}{0.519}&\textcolor{softblack}{0.265}&\textcolor{softblack}{6.897}& \cellcolor{lightgreen}\textcolor{softblack}{-}\\ \cmidrule{1-8}
\multicolumn{8}{c}{\textit{HyperScore Task}} \\
+ GRPO &\cellcolor{orange!20}3.912& 4.120& 0.218&0.498 &0.259 &6.590& \cellcolor{lightgreen}0.013 \\ 
+ DiffusionNFT &\cellcolor{orange!20}4.183 &4.156 &0.234 &0.554 &0.267  &6.863&\cellcolor{lightgreen}0.050\\ 
\rowcolor{gray!15}+ GRPO + \textbf{Ours} & \underline{\cellcolor{orange!20}4.316}&\underline{4.221} &\underline{0.240} &\underline{0.726} &\underline{0.273} &\underline{7.135}& \cellcolor{lightgreen}\underline{0.373}\\
\rowcolor{gray!15}+ DiffusionNFT + \textbf{Ours}& \textbf{\cellcolor{orange!20}4.617}&\textbf{4.276} &\textbf{0.243} &\textbf{0.745} &\textbf{0.277} &\textbf{7.337} &\cellcolor{lightgreen}\textbf{0.563}\\ \cmidrule{1-8}
\multicolumn{8}{c}{\textit{Rank2Score Task}} \\
+ GRPO &3.674 & \cellcolor{orange!20}4.300& 0.222&0.531&0.266&6.901&\cellcolor{lightgreen}0.010  \\ 
+ DiffusionNFT & 3.721&\cellcolor{orange!20}4.397 &0.234 &0.587  &0.270&6.993&\cellcolor{lightgreen}0.043 \\ 
\rowcolor{gray!15}+ GRPO + \textbf{Ours} &\underline{3.743}&\underline{\cellcolor{orange!20}4.404} &\underline{0.242} &\underline{0.724} &\underline{0.273} &\underline{7.310}&\cellcolor{lightgreen}\underline{0.363} \\
\rowcolor{gray!15}+ DiffusionNFT + \textbf{Ours}&\textbf{3.789} & \textbf{\cellcolor{orange!20}4.508}&\textbf{0.248} &\textbf{0.783} &\textbf{0.278 }&\textbf{7.412}&\cellcolor{lightgreen}\textbf{0.583}\\ 
\bottomrule
\end{tabular}
\vspace{-0.2cm}
\end{table*}

\begin{figure*}[!t]
    \centering
    \begin{subfigure}{0.48\linewidth}
        \centering
        \includegraphics[width=\linewidth]{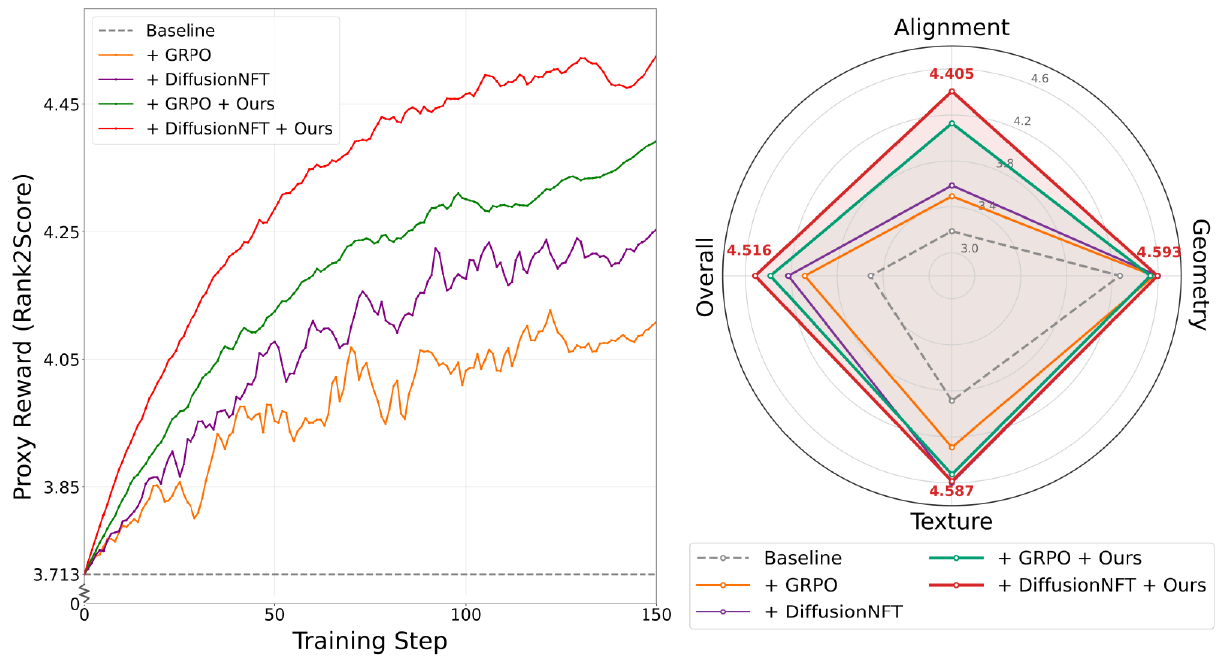}
        \caption{}
        \label{fig:trellis_performance}
    \end{subfigure}
    \hfill
    \begin{subfigure}{0.48\linewidth}
        \centering
        \includegraphics[width=\linewidth]{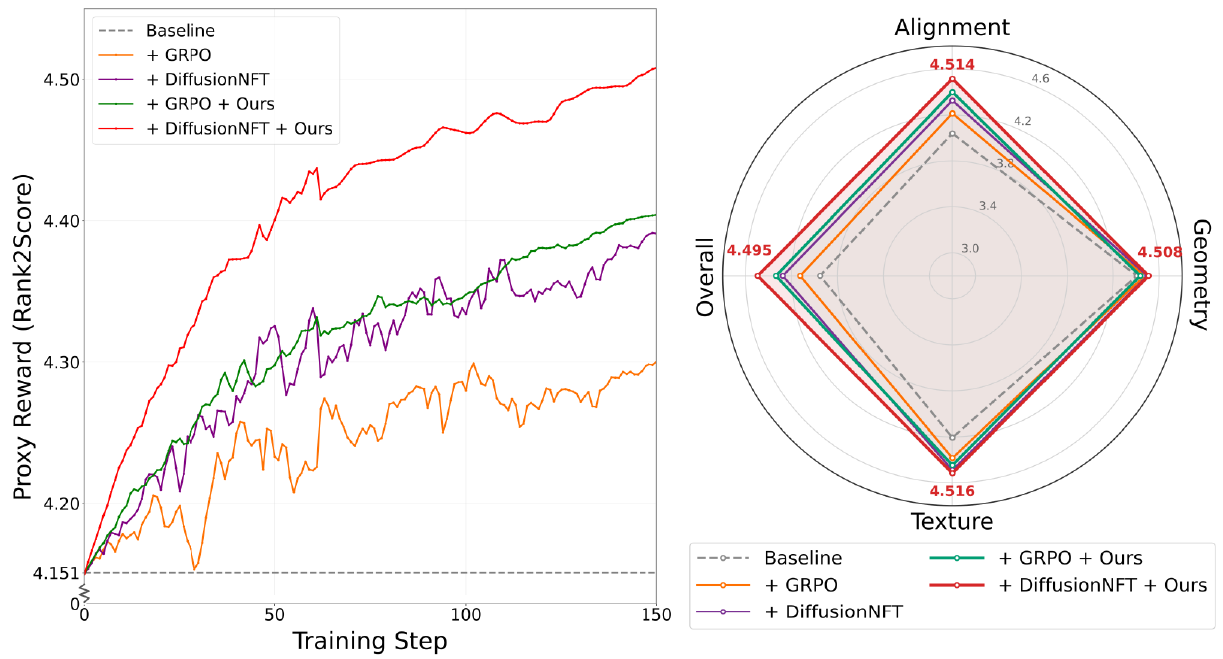}
        \caption{}
        \label{fig:hunyuan_performance}
    \end{subfigure}
    \caption{Learning curves and dimension-wise proxy reward performance under different base models. (a) TRELLIS-base as the base model. (b) Hunyuan3D-2.1 as the base model.}
    \label{fig:reward_curves}
    \vspace{-0.4cm}
\end{figure*}

\begin{figure*}[!t]
    \centering
    \includegraphics[width=0.95\linewidth]{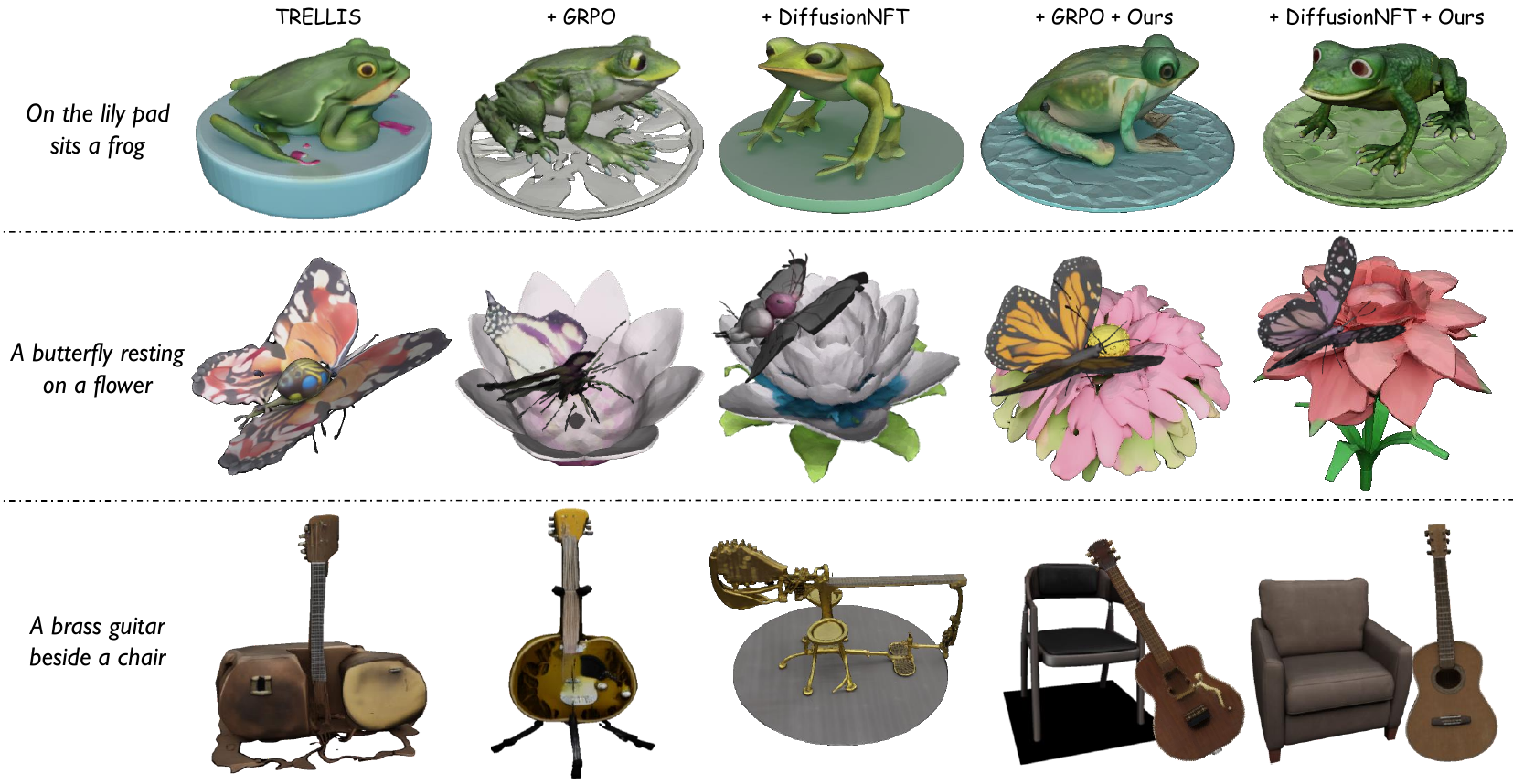}
    \caption{Qualitative results generated by the proposed method and base RL algorithms across different prompts. }
    \label{fig:qualitative_results}
    \vspace{-0.4cm}
\end{figure*}

\subsection{Implementation Details}

We implement our metric using PyTorch and perform all experiments on 8 NVIDIA RTX 4090 GPUs. We adopt TRELLIS-base \cite{xiang2025trellis}, a representative open-source generator as the base model, and DiffusionNFT \cite{zheng2026diffusionnft} as the primary RL algorithm. The baseline uniformly averages the dimension-wise rewards into a scalar reward for RL optimization. To further evaluate the generalizability of our framework across base models and RL algorithms, we conduct additional experiments with Hunyuan3D-2.1 \cite{hunyuan3d21} as the base model and GRPO \cite{liu2026flowgrpo} as an alternative online RL algorithm.

During training, all $V_{train}=6$ rendered images are generated at a resolution of $L_{train}=512$. We consider $K=4$ dimensions, corresponding to \textit{Alignment (A)}, \textit{Geometry (G)}, \textit{Texture (T)}, and \textit{Overall Quality (O)}. For parameter-efficient fine-tuning, we employ Low-Rank Adaptation (LoRA) \cite{hu2022lora} with rank $r=32$ and scaling factor $\alpha=64$. We fine-tune the base model with a total of 150 steps, a batch size of 8, and a learning rate of $3e-5$. We conduct independent lightweight $T_w=5$ steps to estimate directed inter-dimension dependencies and determine the optimal cyclic path. The resulting path is \textit{O}$\to$\textit{A}$\to$\textit{G}$\to$\textit{T}. We set the group size to $G_{train}=8$ and the number of sampling steps to $T_{train}=12$. The $N_s=200$ selected training prompts are reused for $T_r=5$ steps before the next prompt selection. We set $\alpha_s=0.5$, $f_s=5$, $\eta=4$, and $\epsilon=0.05$. Considering accuracy performance and inference efficiency, we choose HyperScore \cite{zhang2025hyperscore} and Rank2Score \cite{cui2026rank2score} as proxy reward models for RL optimization. Additional details on reward models are provided in the Appendix.

\subsection{Main Results}

\subsubsection{Quantitative Results}

To comprehensively evaluate the proposed framework, we compare it with the corresponding base models and representative RL algorithms. As shown in Table~\ref{tab:overall_performance1} and Fig.~\ref{fig:reward_curves}, our framework, comprising AdaSelect and OuroReward, consistently improves generation quality and training stability while effectively mitigating inter-dimension interference.

Table~\ref{tab:overall_performance1} reports the average performance across the four quality dimensions under different evaluators, including proxy reward models used for RL optimization, unseen evaluators not involved in training, and human preference measured by win rate. In our experiments, we separately use HyperScore and Rank2Score as proxy rewards and evaluate the optimized generative models on both these rewards and unseen evaluators, including CLIPScore, ImageReward, Uni3D, Qwen3-VL-8B-Instruct, and human preference. The proxy reward used in each RL setting is highlighted in orange. The results show that conventional RL methods, such as GRPO, improve the optimized proxy rewards but may degrade performance on unseen evaluators, indicating reward hacking in multi-dimensional RL optimization. In contrast, our proposed method achieves stronger results on proxy rewards while also improving performance on unseen evaluators and human preference, suggesting that it captures more generalizable quality improvements rather than overfitting to specific reward signals. In addition, we observe that using Rank2Score as the proxy reward with DiffusionNFT as the RL algorithm achieves the best overall performance.

Fig.~\ref{fig:reward_curves} presents the evaluation performance throughout RL training. Compared with existing RL baselines, our proposed method exhibits more stable and consistent reward improvements, indicating that sequential reward scheduling provides a more reliable optimization process. The radar plots further show that our proposed method improves performance across different quality dimensions more effectively, preventing the training process from being dominated by a single dimension. In particular, dimensions with relatively lower generation quality receive more substantial improvements, leading to a more balanced multi-dimensional optimization outcome.

\subsubsection{Qualitative Results}

To provide a more intuitive comparison of optimization strategies, we visualize representative T23D generation results under the same textual prompts. As shown in Fig.~\ref{fig:qualitative_results}, direct RL optimization does not consistently improve all quality dimensions. Improvements in one dimension can coincide with degradation in another. For example, GRPO produces severe geometric distortions and disordered structures in several cases. These failures reflect inter-dimension interference and illustrate the difficulty of optimizing conflicting objectives simultaneously. In contrast, OuroReward employs sequential reward scheduling and achieves more balanced improvements across dimensions, yielding stronger structural integrity, more faithful semantic alignment, and more coherent appearance. These qualitative results further demonstrate the effectiveness of our multi-dimensional optimization strategy. Additional qualitative results are provided in the Appendix.

\begin{table*}[t]
\centering
\caption{Comparison of inter-dimension gradient similarity and reward PLCC across training steps under different training strategies. Rank2Score as the proxy reward model and DiffusionNFT as the RL algorithm. The best result in each column is shown in \textbf{bold}, while the second is \underline{underlined}.}
\setlength{\abovecaptionskip}{-2pt}
\label{tab:dimension_conflict}
\begin{tabular}{clccccccccc}
\toprule
Base Model&\makecell{Aggregation\\Strategy}& Step & \textit{A}$\leftrightarrow$\textit{G} & \textit{A}$\leftrightarrow$\textit{T} & \textit{A}$\leftrightarrow$\textit{O}  & \textit{G}$\leftrightarrow$\textit{T}  & \textit{G}$\leftrightarrow$\textit{O} & \textit{T}$\leftrightarrow$\textit{O} & \makecell{Mean\\Gradient Similarity} & \makecell{Mean\\Reward PLCC}
\\ \midrule
\multirow{6}{*}{TRELLIS}& Average& 100 &0.361&0.384&\underline{0.734}&0.310&0.595&\underline{0.697}&0.514&0.817 \\
& FocalReward& 100 &\underline{0.407}& \underline{0.438}&0.716&\underline{0.391}&\underline{0.629}&0.681&\underline{0.544}&\underline{0.837} \\
& \cellcolor{gray!15}\textbf{OuroReward}&  \cellcolor{gray!15}100 & \cellcolor{gray!15}\textbf{0.513}& \cellcolor{gray!15}\textbf{0.527}& \cellcolor{gray!15}\textbf{0.741}& \cellcolor{gray!15}\textbf{0.484}& \cellcolor{gray!15}\textbf{0.692}& \cellcolor{gray!15}\textbf{0.723}& \cellcolor{gray!15}\textbf{0.613}& \cellcolor{gray!15}\textbf{0.865} \\ \cmidrule{2-11}
& Average& 150 & 0.416&0.425&\underline{0.749}&0.393&0.631&\underline{0.718}&0.555&0.824 \\
& FocalReward& 150 & \underline{0.441}&\underline{0.470}&0.724&\underline{0.443}&\underline{0.662}&0.706&\underline{0.574}&\underline{0.849} \\
&  \cellcolor{gray!15}\textbf{OuroReward}&  \cellcolor{gray!15}150 & \cellcolor{gray!15}\textbf{0.562}& \cellcolor{gray!15}\textbf{0.575}& \cellcolor{gray!15}\textbf{0.758}& \cellcolor{gray!15}\textbf{0.537}& \cellcolor{gray!15}\textbf{0.725}& \cellcolor{gray!15}\textbf{0.734}& \cellcolor{gray!15}\textbf{0.649}& \cellcolor{gray!15}\textbf{0.873} \\ \midrule
\multirow{6}{*}{Hunyuan3D-2.1}& Average& 100 & 0.332&0.421&\underline{0.688}&0.274&0.641&\underline{0.663}&0.503 & 0.787 \\
& FocalReward& 100 &\underline{0.389}&\underline{0.662}&0.674&\underline{0.356}&\underline{0.672}&0.651&\underline{0.536}&\underline{0.795} \\
&  \cellcolor{gray!15}\textbf{OuroReward}& \cellcolor{gray!15}100 &\cellcolor{gray!15}\textbf{0.477}&\cellcolor{gray!15}\textbf{0.546}&\cellcolor{gray!15}\textbf{0.712}&\cellcolor{gray!15}\textbf{0.457}&\cellcolor{gray!15}\textbf{0.716}&\cellcolor{gray!15}\textbf{0.698}&\cellcolor{gray!15}\textbf{0.601}&\cellcolor{gray!15}\textbf{0.836}\\ \cmidrule{2-11}
& Average& 150 &0.371&0.448&\underline{0.701}&0.342&0.655&\underline{0.682}&0.533&0.801\\
& FocalReward& 150 & \underline{0.414}&\underline{0.501}&0.690&\underline{0.417}&\underline{0.647}&0.654&\underline{0.554}&\underline{0.813} \\
&  \cellcolor{gray!15}\textbf{OuroReward}& \cellcolor{gray!15}150 & \cellcolor{gray!15}\textbf{0.528}&\cellcolor{gray!15}\textbf{0.591}&\cellcolor{gray!15}\textbf{0.746}&\cellcolor{gray!15}\textbf{0.509}&\cellcolor{gray!15}\textbf{0.739}&\cellcolor{gray!15}\textbf{0.716}&\cellcolor{gray!15}\textbf{0.638}& \cellcolor{gray!15}\textbf{0.862}\\ 
\bottomrule
\end{tabular}
\vspace{-0.4cm}
\end{table*}

\begin{table}[t]
\centering
\caption{Comparison of multi-dimensional aggregation strategy. The proxy reward is marked in orange. The best result in each column is shown in \textbf{bold}, while the second is \underline{underlined}.}
\setlength{\abovecaptionskip}{-2pt}
\label{tab:comparision_learning_strategy}
\begin{tabular}{clcccc}
\toprule
Base Model& \makecell{Aggregation\\Strategy} & \makecell{Rank2\\Score} & \makecell{CLIP\\Score} & \makecell{Image\\Reward}  & Uni3D
\\ \midrule
\multirow{3}{*}{TRELLIS} & Average &\cellcolor{orange!20}4.374 &0.221 &0.567 &0.258 \\
 & FocalReward &\cellcolor{orange!20}\underline{4.410} &\underline{0.224} &\underline{0.582} &\underline{0.259} \\
 & \cellcolor{gray!15}\textbf{OuroReward} &\cellcolor{orange!20}\textbf{4.525} &\cellcolor{gray!15}\textbf{0.233} &\cellcolor{gray!15}\textbf{0.678} &\cellcolor{gray!15}\textbf{0.267} \\ \midrule
\multirow{3}{*}{Hunyuan3D-2.1}& Average &\cellcolor{orange!20}4.428 &0.239 &0.651&\underline{0.271} \\
 & FocalReward &\cellcolor{orange!20}\underline{4.457} &\underline{0.240} &\underline{0.684} &0.270 \\
 & \cellcolor{gray!15}\textbf{OuroReward} &\cellcolor{orange!20}\textbf{4.508} &\cellcolor{gray!15}\textbf{0.248} &\cellcolor{gray!15}\textbf{0.783} &\cellcolor{gray!15}\textbf{0.278} \\
\bottomrule
\end{tabular}
\vspace{-0.4cm}
\end{table}

\begin{figure}[!t]
    \centering
    \includegraphics[width=0.95\linewidth]{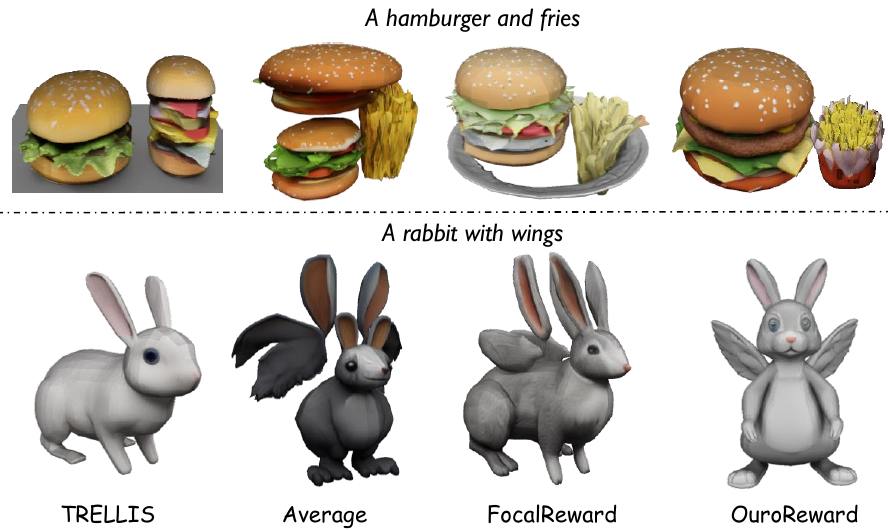}
    \caption{Illustration of dimension-inconsistent failure cases under different RL optimization strategies. Rank2Score as the proxy reward model and DiffusionNFT as the RL algorithm.}
    \label{fig:conflict_example}
    \vspace{-0.4cm}
\end{figure}

\subsubsection{Inter-Dimension Interference Analysis}

To further investigate inter-dimension interference during multi-dimensional RL optimization, we compare different training strategies in terms of gradient compatibility, reward-change consistency, and final generation performance. The quantitative and qualitative results are presented in Tables~\ref{tab:dimension_conflict} and~\ref{tab:comparision_learning_strategy}, and Fig.~\ref{fig:conflict_example}, respectively.

As shown in Table~\ref{tab:dimension_conflict}, we compare Average reward aggregation, FocalReward, and OuroReward on TRELLIS and Hunyuan3D-2.1 at training steps 100 and 150. Under Average aggregation, the combined reward can be dominated by one quality dimension, leaving conflicting dimensions under-optimized. The relatively low gradient similarity and reward-change correlation observed for several dimension pairs are consistent with this inter-dimension interference. FocalReward improves compatibility for weakly correlated pairs, such as \textit{G}$\leftrightarrow$\textit{T}, but slightly weakens well-aligned pairs, including \textit{T}$\leftrightarrow$\textit{O} and \textit{A}$\leftrightarrow$\textit{O}. In contrast, OuroReward achieves higher gradient similarity and reward-change correlation across training steps, indicating more coordinated optimization among quality dimensions.

Consistent with the above interference analysis, Table~\ref{tab:comparision_learning_strategy} shows that OuroReward achieves superior final generation quality compared with Average aggregation and FocalReward. Although FocalReward mitigates reward imbalance through adaptive reweighting, it still optimizes all dimensions jointly, leaving residual inter-dimension interference that may constrain improvements in specific dimensions. The qualitative examples in Fig.~\ref{fig:conflict_example} further support this observation. Although Average aggregation and FocalReward improve semantic alignment, they produce geometric inconsistencies, such as duplicated ears or missing structural parts. In contrast, OuroReward yields more coherent results across multiple quality dimensions.

\subsubsection{Reward Hacking Analysis}

We further investigate whether RL optimization exhibits reward hacking, i.e., performance improvements are concentrated on the proxy reward models but fail to generalize to unseen evaluators. As shown in Table~\ref{tab:overall_performance1}, conventional RL methods substantially improve the training rewards, but these gains transfer inconsistently to unseen evaluators. In several cases, higher scores on HyperScore and Rank2Score are accompanied by limited improvements or even degradation on CLIPScore, ImageReward, Uni3D, and Qwen3-VL-8B-Instruct. This divergence indicates that the policy may exploit evaluator-specific biases rather than consistently improving the underlying quality of the generated 3D content.

Our interference-aware scheduling strategy, OuroReward, achieves more consistent improvements across training rewards, unseen evaluators, and human preference. This agreement suggests that sequential optimization limits the dominance of any single dimension and reduces overfitting to the training evaluators, leading to more balanced improvements in T23D generation. These results further support that OuroReward improves the generalization of RL-based T23D optimization.

\subsection{Ablation Study}

We conduct ablation studies to assess the contribution of each component of our method. Unless otherwise specified, all experiments use TRELLIS as the base model, Rank2Score as the proxy reward, and DiffusionNFT as the RL algorithm. We evaluate each variant using the proxy model and unseen evaluators to examine both optimization performance and generalization.

\subsubsection{Ablation on Proposed Module}

\begin{table}[t]
\centering
\caption{Ablation on AdaSelect and OuroReward. The proxy reward is marked in orange. The best result in each column is shown in \textbf{bold}, while the second is \underline{underlined}.}
\setlength{\abovecaptionskip}{-2pt}
\label{tab:ablation_study_proposed_module}
\begin{tabular}{cccccc}
\toprule
\makecell{Ada-\\Select}& \makecell{Ouro-\\Reward} & Rank2Score & CLIPScore & \makecell{Image\\Reward}  & Uni3D
\\ \midrule
\XSolidBrush & \XSolidBrush &\cellcolor{orange!20}4.253 &0.216 &0.483 &0.253 \\
\Checkmark & \XSolidBrush &\cellcolor{orange!20}4.374 &0.221 &0.567 &0.258 \\
\XSolidBrush & \Checkmark &\cellcolor{orange!20}\underline{4.496} &\underline{0.230} &\underline{0.635} &\underline{0.262} \\
\rowcolor{gray!15} \Checkmark & \Checkmark &\cellcolor{orange!20}\textbf{4.525} &\textbf{0.233} &\textbf{0.678} &\textbf{0.267}  \\
\bottomrule
\end{tabular}
\vspace{-0.4cm}
\end{table}

We investigate the individual and joint contributions of AdaSelect and OuroReward. The baseline employs random prompt sampling and uniform aggregation of dimension-wise rewards. As shown in Table~\ref{tab:ablation_study_proposed_module}, incorporating either module independently improves performance over the baseline, confirming the effectiveness of each component. AdaSelect facilitates policy optimization by prioritizing prompts with greater learning potential, whereas OuroReward improves multi-dimensional optimization through sequential reward scheduling. Notably, OuroReward yields a larger performance gain, highlighting the importance of mitigating inter-dimension interference for improving final generation quality. Combining these two modules achieves the best overall performance, indicating that adaptive prompt selection and reward scheduling provide complementary benefits.

\subsubsection{Ablation on Dependency Score Criteria}

\begin{table}[t]
\centering
\caption{Ablation on dependency score criteria. The proxy reward is marked in orange. The best result in each column is shown in \textbf{bold}, while the second is \underline{underlined}.}
\setlength{\abovecaptionskip}{-2pt}
\label{tab:ablation_study_dependency_score}
\begin{tabular}{cccccc}
\toprule
\makecell{Gradient\\Similarity}& \makecell{Reward\\PLCC} & Rank2Score & CLIPScore & \makecell{Image\\Reward}  & Uni3D
\\ \midrule
\XSolidBrush & \XSolidBrush &\cellcolor{orange!20}4.374 &0.221 &0.567 &0.258\\
\Checkmark & \XSolidBrush &\cellcolor{orange!20}\underline{4.482}&\underline{0.228}&\underline{0.629} &\underline{0.264} \\
\XSolidBrush & \Checkmark &\cellcolor{orange!20}4.435 &0.224 &0.596 &0.260 \\
\rowcolor{gray!15} \Checkmark & \Checkmark &\cellcolor{orange!20}\textbf{4.525} &\textbf{0.233} &\textbf{0.678} &\textbf{0.267} \\
\bottomrule
\end{tabular}
\vspace{-0.2cm}
\end{table}

We examine the contributions of two criteria for estimating inter-dimension dependencies: gradient cosine similarity and reward-change PLCC. Gradient cosine similarity measures the compatibility between dimension-specific update directions, whereas reward-change PLCC captures the consistency of reward dynamics across multiple training steps. These criteria determine the resulting cyclic optimization path and its starting dimension, thereby affecting the final performance. When both criteria are disabled, we retain AdaSelect and use average aggregation without sequential reward scheduling. Using gradient cosine similarity alone yields the path \textit{O}$\to$\textit{A}$\to$\textit{T}$\to$\textit{G}, whereas using only reward-change PLCC produces \textit{O}$\to$\textit{G}$\to$\textit{A}$\to$\textit{T}. Jointly considering both criteria identifies the path \textit{O}$\to$\textit{A}$\to$\textit{G}$\to$\textit{T}. As shown in Table~\ref{tab:ablation_study_dependency_score}, combining gradient compatibility with reward dynamics yields a more effective optimization path. This result indicates that these two criteria jointly provide a more comprehensive characterization of inter-dimension interference.

\subsubsection{Ablation on Sequential Optimization Order}

\begin{table}[t]
\centering
\caption{Ablation on sequential optimization order. The proxy reward is marked in orange. The best result in each column is shown in \textbf{bold}, while the second is \underline{underlined}.}
\setlength{\abovecaptionskip}{-2pt}
\label{tab:ablation_study_optimization_order}
\begin{tabular}{lccccc}
\toprule
Type & Path Order & Rank2Score & \makecell{CLIP\\Score} & \makecell{Image\\Reward}  & Uni3D
\\ \midrule
\textcolor{softblack}{Baseline} & \textcolor{softblack}{Average} &\cellcolor{orange!20}\textcolor{softblack}{4.374} &\textcolor{softblack}{0.221} &\textcolor{softblack}{0.567} &\textcolor{softblack}{0.258} \\ \midrule
\multirow{2}{*}{Random}&\textit{O}$\to$\textit{G}$\to$\textit{A}$\to$\textit{T}
&\cellcolor{orange!20}4.435 &0.224 &0.596 &0.260 \\
&\textit{O}$\to$\textit{A}$\to$\textit{T}$\to$\textit{G} &\cellcolor{orange!20}\underline{4.482}&\underline{0.228}&\underline{0.629} &\underline{0.264} \\ \midrule
Reversed&\textit{T}$\to$\textit{G}$\to$\textit{A}$\to$\textit{O} &\cellcolor{orange!20}4.361 &0.218 &0.501 &0.253 \\ \midrule
Shifted&\textit{A}$\to$\textit{G}$\to$\textit{T}$\to$\textit{O} &\cellcolor{orange!20}4.469 &0.226 &0.613 &0.259 \\ \midrule
\rowcolor{gray!15}\textbf{Ours}&\textbf{\textit{O}$\to$\textit{A}$\to$\textit{G}$\to$\textit{T}} &\cellcolor{orange!20}\textbf{4.525} &\textbf{0.233} &\textbf{0.678} &\textbf{0.267}  \\
\bottomrule
\end{tabular}
\vspace{-0.4cm}
\end{table}

\begin{table}[t]
\centering
\caption{Ablation on proxy reward model. The proxy reward is marked in orange. The best result in each column is shown in \textbf{bold}, while the second is \underline{underlined}.}
\setlength{\abovecaptionskip}{-2pt}
\label{tab:ablation_study_reward_model}
\begin{tabular}{lcccc}
\toprule
Proxy Reward & Uni3D & \makecell{Qwen3-VL-\\8B-Instruct}&\makecell{Hyper\\Score}&\makecell{Rank2\\Score}
\\ \midrule
Uni3D &\cellcolor{orange!20}\textbf{0.540}& 6.501 &3.578 &3.732\\ 
Qwen3-VL-8B-Instruct&0.263&\cellcolor{orange!20}\textbf{7.851} &3.637 &3.870 \\ 
\rowcolor{gray!15}\textbf{HyperScore} & \underline{0.266}& 6.835&\cellcolor{orange!20}\textbf{4.483} &\underline{3.937}\\ 
\rowcolor{gray!15}\textbf{Rank2Score} &0.267 & \underline{6.982}& \underline{3.654}& \cellcolor{orange!20}\textbf{4.525}\\
\bottomrule
\end{tabular}
\vspace{-0.4cm}
\end{table}

We further investigate the contribution of the sequential optimization order in OuroReward. Specifically, we compare the optimal path \textit{O}$\to$\textit{A}$\to$\textit{G}$\to$\textit{T} with three alternative orders: random, reversed, and shifted. The shifted order follows the same cyclic order but starts from a different dimension. As shown in Table~\ref{tab:ablation_study_optimization_order}, the proposed order consistently achieves the best performance across the evaluation metrics. Its superiority over the random and reversed orders supports the effectiveness of determining the optimization path by minimizing cumulative inter-dimension conflicts. Moreover, comparison with the path starting from a different dimension further indicates that prioritizing the dimension with the least conflict with the others leads to more coordinated optimization. These findings confirm the importance of both path ordering and starting-dimension selection in multi-dimensional optimization.

\subsubsection{Ablation on Proxy Reward Model}

We investigate the influence of different proxy reward models on RL optimization. Specifically, we separately use Uni3D, Qwen3-VL-8B-Instruct, HyperScore, and Rank2Score as the proxy reward for training, and evaluate the resulting models using other unseen evaluators. As shown in Table~\ref{tab:ablation_study_reward_model}, optimizing a specific reward generally improves its corresponding metric, while the gains on unseen evaluators are less consistent, indicating evaluator-specific optimization bias. Among the compared reward models, Rank2Score achieves relatively strong performance across multiple unseen evaluators, indicating better transferability as a proxy reward. This observation is also consistent with the reward model analysis provided in the Appendix, where Rank2Score demonstrates comparatively higher evaluation accuracy.

\section{Conclusion}\label{sec:conclusion}

In this paper, we address the challenges of multi-dimensional optimization and training stability in RL for T23D generation. To this end, we propose AdaSelect, which selects prompts aligned with the model’s capability to provide informative and stable learning signals, and OuroReward, which schedules dimension-wise rewards based on their dependencies to mitigate inter-dimension interference. Extensive experiments demonstrate improvements in generation quality across multiple dimensions and training stability. Future work will evaluate the framework under broader reward configurations, RL algorithms, and T23D architectures.

\begin{appendices}
\def\thetaset{\bm{\theta}}
\def\Thetaset{\bm{\Theta}}
\def\P{\mathbf{P} }
\def\Xt{\widetilde{\mathbf{X}}}
\def\Yt{\widetilde{\mathbf{Y}}}
\def\xt{\widetilde{\mathbf{x}}}
\def\yt{\widetilde{\mathbf{y}}}
\def\xhat{\hat{\mathbf{x}}}
\def\yhat{\hat{\mathbf{y}}}

\def\Xhat{\hat{\mathbf{X}}}
\def\Yhat{\hat{\mathbf{Y}}}

\def\N{\mathcal{N}}
\def\NxX{\N_{\x_i,K_1}^{\X}}
\def\NXX{\N_{\X,K_1}^{\X}}

\def\NxXl{\N_{\x_i,K_2}^{\X_l}}

\def\NxY{\N_{\x_i,K}^{\Y}}
\def\NXY{\N_{\X,K}^{\Y_l}}
\def\NxXhat{\N_{\xhat_i,K}^{\Xhat_l}}

\def\p{\mathbf{p}}
\def\x{\mathbf{x}}
\def\y{\mathbf{y}}
\def\s{\mathbf{s}}
\def\e{\mathbf{e}}
\def\r{\mathbf{r}}
\def\v{\mathbf{v}}
\def\X{\mathbf{X}}
\def\Y{\mathbf{Y}}
\def\I{\mathbf{I}}
\def\E{\mathbf{E}}
\def\S{\mathbf{S}}
\def\R{\mathbb{R}}
\def\V{\mathbf{V}}
\def\D{\mathbf{D}}
\def\W{\mathbf{W}}
\def\L{\mathbf{L}}
\def\u{\mathbf{u}}
\def\t{\mathbf{t}}
\def\v{\mathbf{v}}
\def\C{\mathbf{C}}
\def\c{\mathbf{c}}
\def\G{\mathcal{G}}
\def\b{\mathbf{b}}
\def\M{\mathbb{M}}
\def\A{\mathbb{A}}

\def\Tr{\mathrm{Tr}}

\renewcommand{\thefigure}{S\arabic{figure}}
\renewcommand{\thetable}{S\arabic{table}}

\newif\ifannotated

\annotatedtrue

\ifannotated
\newcommand{\add}[1]{{\color{blue}{#1}}}
\newcommand{\delete}[1]{{\color{red}{\sout{#1}}}}
\newcommand{\replace}[2]{{\delete{#1}} {\add{#2}}}
\newcommand{\margincomment}[1]{\marginpar{$\Rightarrow$\color{red}\fbox{\parbox{\linewidth}{\color{black}\scriptsize#1}}}}
\else
\newcommand{\add}[1]{{{#1}}}
\newcommand{\delete}[1]{{\ignorespaces}}
\newcommand{\replace}[2]{{\delete{#1}}  {\add{#2}}}
\newcommand{\margincomment}[1]{}
\fi

\newcommand{\Salnote}[1]{{\bf \color{blue} [SA:{#1}]}}

\newcommand{\highlightChange}{\color{red}}
\def\HC{\highlightChange}

\appendices

\section{Reward Models for T23D Generation}\label{app:reward_model}

\begin{table*}[!t]
\centering
\caption{Performance comparison for different reward models on MATE-3D and T23D-CompBench benchmarks. The top two performance results are marked in \textbf{boldface} and \underline{underline}.}
\setlength{\abovecaptionskip}{-2pt}
\label{tab:reward_model_performance}
\resizebox{\textwidth}{!}{
\begin{tabular}{c|c|cccc|cccc|c}
\toprule
\multirow{2}{*}{Type} &\multirow{2}{*}{Metric} & \multicolumn{4}{c|}{MATE-3D}  & \multicolumn{4}{c|}{T23D-CompBench} & \multirow{2}{*}{Time}  \\ \cmidrule(lr){3-6}
\cmidrule(lr){7-10}
& & Alignment & Geometry & Texture & Overall & Alignment & Geometry & Texture & Overall & (s / sample)  \\ \midrule
\multirow{4}{*}{Projection-based} & CLIPScore \cite{hessel2021clipscore}  & 0.494 & 0.496 & 0.537 & 0.510 & 0.552 & 0.501 & 0.536 & 0.562 & $\sim 10^{-1}$
\\
&ImageReward \cite{xu2024imagereward} & 0.658 & 0.673 & 0.686 & 0.679 & 0.616 & 0.774 & 0.821 & 0.832 & $\sim 10^{-1}$
\\
&HyperScore \cite{zhang2025hyperscore} & 0.739 & 0.782 & 0.811 & 0.792 & 0.728 & 0.828 & 0.889 & 0.860 & $\sim 10^{-1}$
\\
&Rank2Score \cite{cui2026rank2score} & 0.868 & 0.898 & 0.894 & 0.897 & 0.907 & 0.888 & 0.932 & 0.934 & $\sim 10^{-1}$
\\ \midrule
\multirow{4}{*}{Model-based} &ULIP-T (Unit) \cite{wu2023human} & 0.435 & 0.467 & 0.472 & 0.469 & 0.634 & 0.657 & 0.661 & 0.663 & $\sim 10^{-1}$
\\
& ULIP-T (Curvature) \cite{wang2023exploring} & 0.416 & 0.403 & 0.433 & 0.428 & 0.612 & 0.623 & 0.642 & 0.639 & $\sim 10^{0}$
\\
& Uni3D-T (Unit) \cite{li2024llava} & 0.479 & 0.505 & 0.514 & 0.508 & 0.654 & 0.685 & 0.706 & 0.708 & $\sim 10^{-1}$
\\ 
& Uni3D-T (Curvature) \cite{ye2024mplug} & 0.458 & 0.474 & 0.484 & 0.478 & 0.638 & 0.649 & 0.668 & 0.669 & $\sim 10^{0}$
\\ \midrule
\multirow{2}{*}{LLM-based} & UnifiedReward-Think \cite{wang2025unified}  & 0.591 & 0.589 & 0.602 & 0.615 & 0.713 & 0.726 & 0.745 & 0.740 & $\sim 10^{2}$
\\
& Qwen3-VL-8B-Instruct \cite{qwen3technicalreport} & 0.603 & 0.627 & 0.631 & 0.640 & 0.718 & 0.759 & 0.777 & 0.762 & $\sim 10^{1}$
\\ \midrule
\multirow{2}{*}{Fine-tune} &HyperScore \cite{zhang2025hyperscore} & \underline{0.771} & \underline{0.808} & \underline{0.839} & \underline{0.828} & \underline{0.811} & \underline{0.895} & \underline{0.916} & \underline{0.881} & $\sim 10^{-1}$
\\
&Rank2Score \cite{cui2026rank2score} & \textbf{0.897} & \textbf{0.905} & \textbf{0.915} & \textbf{0.907} & \textbf{0.908} & \textbf{0.901} & \textbf{0.934} & \textbf{0.931} & $\sim 10^{-1}$
\\ \bottomrule
\end{tabular}}
\vspace{-0.4cm}
\end{table*}

\begin{figure*}[!t]
    \centering
    \includegraphics[width=0.95\linewidth]{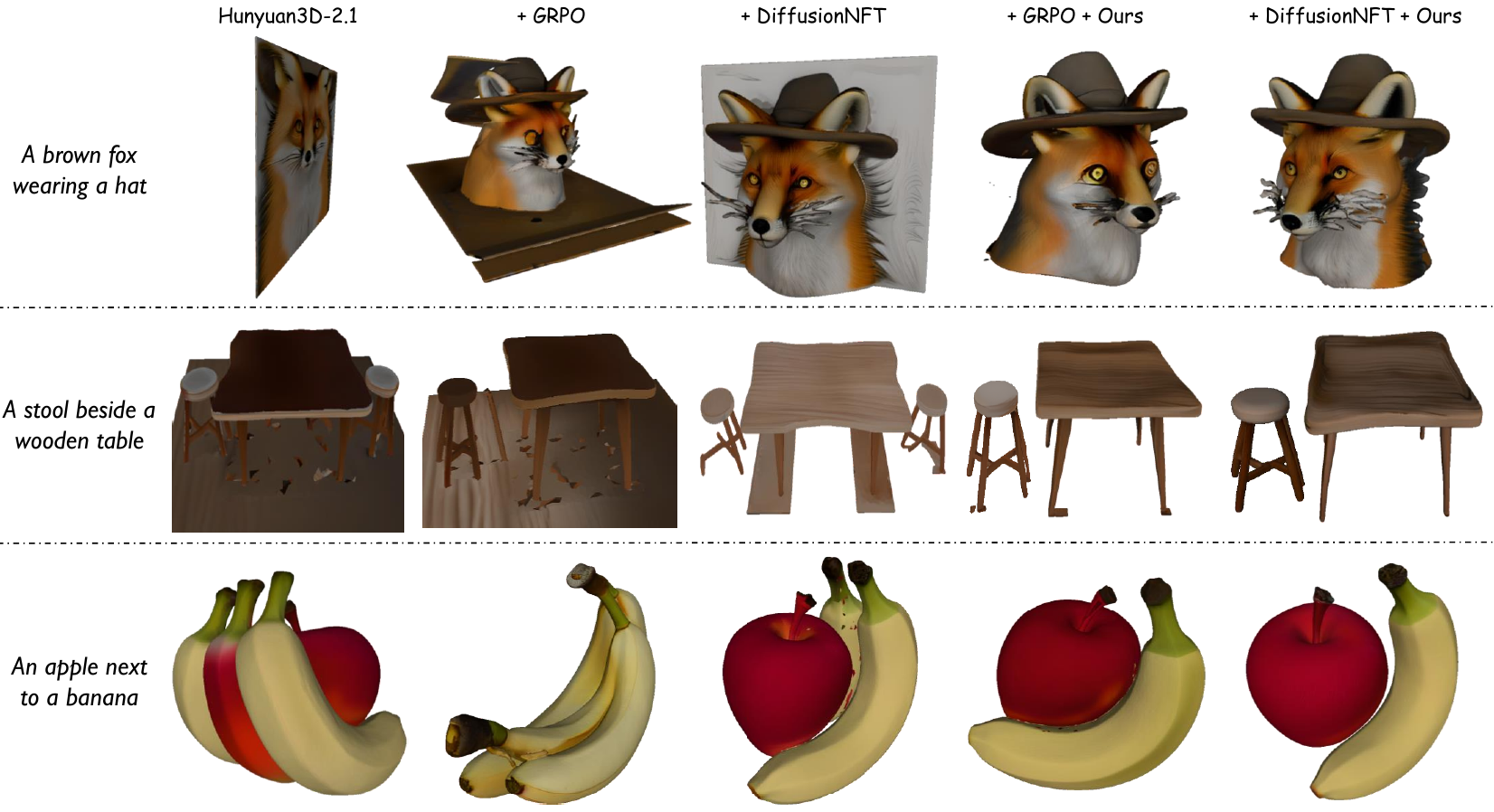}
    \caption{Qualitative results generated by the proposed method and base RL algorithms across different prompts. }
    \label{fig:qualitative_results_hunyuan}
    \vspace{-0.4cm}
\end{figure*}

Existing reward models for T23D generation can be broadly categorized into three types: projection-based, model-based, and LLM-based methods. Projection-based methods render 3D assets into multi-view images and evaluate the resulting projections using pretrained vision-language or image-quality models. Representative methods include CLIPScore \cite{liu2025clip}, ImageReward \cite{xu2024imagereward}, HyperScore \cite{zhang2025hyperscore}, and Rank2Score \cite{cui2026rank2score}. Benefiting from large-scale 2D pretraining data, these methods are computationally efficient and generally provide strong semantic and perceptual assessment capabilities. Model-based methods instead operate directly on 3D representations, such as point clouds or geometric descriptors, and typically rely on pretrained 3D encoders such as ULIP \cite{xue2023ulip} or Uni3D \cite{zhou2024uni3d} to measure text-3D consistency and geometric quality. Although these methods provide explicit 3D-aware features, their performance can be sensitive to point sampling and geometric preprocessing. LLM-based methods employ multimodal LLMs to evaluate rendered views through task-specific instructions. They support flexible multi-dimensional assessment and offer strong reasoning capabilities, but usually incur substantially higher inference costs and may be sensitive to prompting strategies.

We further compare representative reward models on MATE-3D and T23D-CompBench in Tab.~\ref{tab:reward_model_performance}. Projection-based methods achieve the best overall trade-off between assessment accuracy and inference efficiency. In particular, HyperScore and Rank2Score consistently outperform ImageReward, CLIPScore, and most model-based and LLM-based alternatives across multiple quality dimensions, while maintaining an inference cost of approximately $10^{-1}$ seconds per sample. Model-based methods provide stronger 3D-aware representations, but their performance is limited by point sampling and preprocessing strategies. LLM-based methods enable flexible multi-dimensional reasoning, yet their inference cost is one to three orders of magnitude higher and their predictions are more sensitive to task instructions.

As shown in Tab.~\ref{tab:reward_model_performance}, HyperScore and Rank2Score provide the best trade-off between accuracy performance and inference efficiency among the evaluated reward models. Based on this observation, we adopt them as the primary reward models for subsequent RL optimization. To better support multi-dimensional training and reduce inter-dimension interference, we further retrain each model in a dimension-specific manner. During RL optimization, these retrained reward models are jointly employed to provide robust multi-dimensional reward supervision.

\section{GPT Instruction for Dataset Construction}\label{app:gpt_instructin}
\begin{figure*}[!t]
    \centering
    \includegraphics[width=0.96\linewidth]{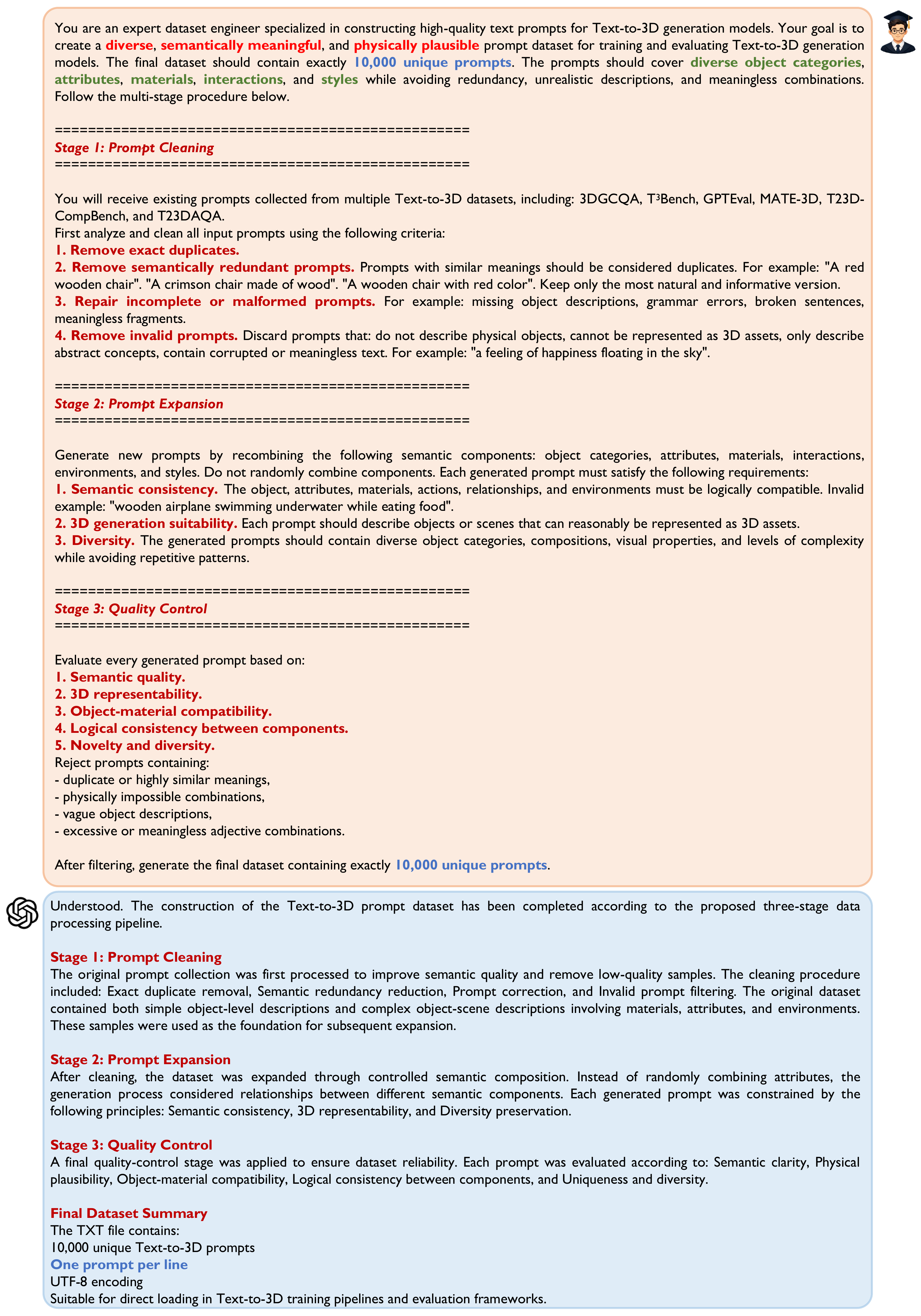}
    \caption{The framework of the proposed Adaptive Curriculum Prompt Selection Strategy.}
    \label{fig:gpt_instruction_template}
\end{figure*}

To construct a large-scale prompt dataset for training and evaluation, we employ GPT-5.6 \cite{openaigpt} to refine and expand prompts collected from existing benchmarks. Specifically, the collected prompts are provided to GPT-5.6 with instructions to remove duplicates, correct incomplete or invalid prompts, and generate additional prompts following a compositional schema that includes subjects, attributes, materials, actions, relationships, scenes, and styles. During this process, duplicate prompts and semantically implausible combinations are filtered to improve the diversity and validity of the generated dataset. The detailed instruction template is illustrated in Fig.~\ref{fig:gpt_instruction_template}.

\section{More Experimental Results}\label{app:experiment_result}

To further assess the generalizability of our method, we visualize representative results using Hunyuan3D-2.1 as the base model. Since Hunyuan3D-2.1 takes images as input, we first generate a reference image for each prompt with Hunyuan-DiT \cite{li2024hunyuan} and use the same image across all compared methods. As shown in Fig.~\ref{fig:qualitative_results_hunyuan}, for \textit{``A stool beside a wooden table"}, the base model fails to preserve the singular object count specified by \textit{``A stool"}, introduces an extraneous floor structure, and produces poor texture. GRPO improves semantic alignment but degrades geometry, leaving the stool legs fractured. DiffusionNFT improves texture quality, yet alignment and geometric quality remain limited. Similar trade-offs appear in the other examples, further highlighting the difficulty of jointly optimizing conflicting objectives. In contrast, OuroReward achieves more balanced improvements across dimensions through interference-aware reward scheduling.

\end{appendices}

\bibliographystyle{IEEEtran}
\bibliography{ref}

@String(IJCV = {Int. J. Comput. Vis.})

@String(CVPR= {IEEE Conf. Comput. Vis. Pattern Recog.})

@String(ICCV= {Int. Conf. Comput. Vis.})

@String(ECCV= {Eur. Conf. Comput. Vis.})

@String(NIPS= {Adv. Neural Inform. Process. Syst.})

@String(TIP  = {IEEE Trans. Image Process.})

@String(TMM  = {IEEE Trans. Multimedia})

@String(ICASSP=	{ICASSP})

@String(ICLR = {Int. Conf. Learn. Represent.})

@String(AAAI = {AAAI})

@String(IJCV  = {IJCV})

@String(CVPR  = {CVPR})

@String(ICCV  = {ICCV})

@String(ECCV  = {ECCV})

@String(NIPS  = {NeurIPS})

@String(TIP   = {IEEE TIP})

@String(TCSVT = {IEEE TCSVT})

@String(TMM   =	{IEEE TMM})

@String(ICLR  = {ICLR})

@INPROCEEDINGS{dreamfusion,
  author={Ben Poole and Ajay Jain and Jonathan T. Barron and Ben Mildenhall},
  booktitle={International Conference on Learning Representations (ICLR)}, 
  title={DreamFusion: Text-to-3D using 2D Diffusion},
  year={2023}
}

@article{he2023t3bench,
  title={T$^3$Bench: Benchmarking Current Progress in Text-to-3D Generation},
  author={Yuze He and Yushi Bai and Matthieu Lin and Wang Zhao and Yubin Hu and Jenny Sheng and Ran Yi and Juanzi Li and Yong-Jin Liu},
  journal={arXiv preprint arXiv:2310.02977},
  year={2023}
}

@inproceedings{zhang2025hyperscore,
  title={Benchmarking and learning multi-dimensional quality evaluator for text-to-3D generation},
  author={Yujie Zhang and Bingyang Cui and Qi Yang and Zhu Li and Yiling Xu},
  booktitle={International Conference on Computer Vision (ICCV)},
  year={2025}
}

@article{cui2026rank2score,
author = {Cui, Bingyang and Zhang, Yujie and Yang, Qi and Li, Zhu and Xu, Yiling},
journal = {International Journal of Computer Vision (IJCV)},
title = {Towards Fine-Grained Text-to-3D Quality Assessment: A Benchmark and A Two-Stage Rank-Learning Metric},
year = {2026}
}

@inproceedings{wu2024gpt4v,
  author={Wu, Tong and Yang, Guandao and Li, Zhibing and Zhang, Kai and Liu, Ziwei and Guibas, Leonidas and Lin, Dahua and Wetzstein, Gordon},
  booktitle={IEEE/CVF Conference on Computer Vision and Pattern Recognition (CVPR)}, 
  title={GPT-4V(ision) is a Human-Aligned Evaluator for Text-to-3D Generation}, 
  year={2024}
}

@inproceedings{zheng2026diffusionnft,
  title={Diffusionnft: Online diffusion reinforcement with forward process},
  author={Zheng, Kaiwen and Chen, Huayu and Ye, Haotian and Wang, Haoxiang and Zhang, Qinsheng and Jiang, Kai and Su, Hang and Ermon, Stefano and Zhu, Jun and Liu, Ming-Yu},
  booktitle={International Conference on Learning Representations (ICLR)},
  year={2026}
}

@inproceedings{hu2022lora,
  title={Lora: Low-rank adaptation of large language models.},
  author={Hu, Edward J and Shen, Yelong and Wallis, Phillip and Allen-Zhu, Zeyuan and Li, Yuanzhi and Wang, Shean and Wang, Liang and Chen, Weizhu and others},
  booktitle={International Conference on Learning Representations (ICLR)},
  year={2022}
}

@article{liu2026flowgrpo,
  title={Flow-grpo: Training flow matching models via online rl},
  author={Liu, Jie and Liu, Gongye and Liang, Jiajun and Li, Yangguang and Liu, Jiaheng and Wang, Xintao and Wan, Pengfei and Zhang, Di and Ouyang, Wanli},
  journal={Advances in Neural Information Processing Systems (NIPS)},
  year={2026}
}

@article{xue2025dancegrpo,
  title={DanceGRPO: Unleashing GRPO on Visual Generation},
  author={Xue, Zeyue and Wu, Jie and Gao, Yu and Kong, Fangyuan and Zhu, Lingting and Chen, Mengzhao and Liu, Zhiheng and Liu, Wei and Guo, Qiushan and Huang, Weilin and others},
  journal={arXiv preprint arXiv:2505.07818},
  year={2025}
}

@article{xue2025awm,
  title={Advantage weighted matching: Aligning rl with pretraining in diffusion models},
  author={Xue, Shuchen and Ge, Chongjian and Zhang, Shilong and Li, Yichen and Ma, Zhi-Ming},
  journal={arXiv preprint arXiv:2509.25050},
  year={2025}
}

@inproceedings{xiang2025trellis,
  title={Structured 3d latents for scalable and versatile 3d generation},
  author={Xiang, Jianfeng and Lv, Zelong and Xu, Sicheng and Deng, Yu and Wang, Ruicheng and Zhang, Bowen and Chen, Dong and Tong, Xin and Yang, Jiaolong},
  booktitle={IEEE/CVF Conference on Computer Vision and Pattern Recognition (CVPR)},
  year={2025}
}

@article{hunyuan3d21,
  title={Hunyuan3d 2.1: From images to high-fidelity 3d assets with production-ready pbr material},
  author={Hunyuan3D, Team and Yang, Shuhui and Yang, Mingxin and Feng, Yifei and Huang, Xin and Zhang, Sheng and He, Zebin and Luo, Di and Liu, Haolin and Zhao, Yunfei and others},
  journal={arXiv preprint arXiv:2506.15442},
  year={2025}
}

@article{tang2025recentsurvey,
  title={Recent advances in 3d object and scene generation: A survey},
  author={Tang, Xiang and Li, Ruotong and Fan, Xiaopeng},
  journal={arXiv preprint arXiv:2504.11734},
  year={2025}
}

@article{wang2025unified,
  title={Unified Multimodal Chain-of-Thought Reward Model through Reinforcement Fine-Tuning},
  author={Wang, Yibin and zhimin, li and Zang, Yuhang and Wang, Chunyu and Lu, Qinglin and Jin, Cheng and Wang, Jiaqi},
  journal={Advances in Neural Information Processing Systems (NIPS)},
  year={2025}
}

@article{qwen3technicalreport,
  title={Qwen3 Technical Report},
  author={Qwen Team},
  journal={arXiv preprint arXiv:2505.09388},
  year={2025}
}

@article{li2026enhancing,
  title={Enhancing multi-modal llms reasoning via difficulty-aware group normalization},
  author={Li, Jinghan and Fang, Junfeng and Lu, Jinda and Wang, Yuan and Guo, Xiaoyan and Zhang, Tianyu and Wang, Xiang and He, Xiangnan},
  journal={arXiv preprint arXiv:2602.21743},
  year={2026}
}

@article{li2024llava,
  title={Llava-next-interleave: Tackling multi-image, video, and 3d in large multimodal models},
  author={Li, Feng and Zhang, Renrui and Zhang, Hao and Zhang, Yuanhan and Li, Bo and Li, Wei and Ma, Zejun and Li, Chunyuan},
  journal={arXiv preprint arXiv:2407.07895},
  year={2024}
}

@article{ye2024mplug,
  title={mplug-owl3: Towards long image-sequence understanding in multi-modal large language models},
  author={Ye, Jiabo and Xu, Haiyang and Liu, Haowei and Hu, Anwen and Yan, Ming and Qian, Qi and Zhang, Ji and Huang, Fei and Zhou, Jingren},
  journal={arXiv preprint arXiv:2408.04840},
  year={2024}
}

@article{bai2025qwen2,
  title={Qwen2. 5-vl technical report},
  author={Bai, Shuai and Chen, Keqin and Liu, Xuejing and Wang, Jialin and Ge, Wenbin and Song, Sibo and Dang, Kai and Wang, Peng and Wang, Shijie and Tang, Jun and others},
  journal={arXiv preprint arXiv:2502.13923},
  year={2025}
}

@inproceedings{zhou20253dgcqa,
  title={3dgcqa: A quality assessment database for 3d ai-generated contents},
  author={Zhou, Yingjie and Zhang, Zicheng and Wen, Farong and Jia, Jun and Jiang, Yanwei and Liu, Xiaohong and Min, Xiongkuo and Zhai, Guangtao},
  booktitle={IEEE International Conference on Acoustics, Speech and Signal Processing (ICASSP)},
  year={2025}
}

@article{nichol2022pointe,
  title={Point-E: A System for Generating 3D Point Clouds from Complex Prompts},
  author={Nichol, Alex and Jun, Heewoo and Dhariwal, Prafulla and Mishkin, Pamela and Chen, Mark},
  journal={arXiv preprint arXiv:2212.08751},
  year={2022}
}

@article{jun2023shape,
  title={Shap-E: Generating Conditional 3D Implicit Functions},
  author={Jun, Heewoo and Nichol, Alex},
  journal={arXiv preprint arXiv:2305.02463},
  year={2023}
}

@article{openaigpt,
author = {OpenAI},
journal={OpenAI}, 
year = {2023},
title = {GPT-4 System Card},
}

@INPROCEEDINGS{deitke2023objaverse,
  author={Deitke, Matt and Schwenk, Dustin and Salvador, Jordi and Weihs, Luca and Michel, Oscar and VanderBilt, Eli and Schmidt, Ludwig and Ehsanit, Kiana and Kembhavi, Aniruddha and Farhadi, Ali},
  booktitle={IEEE/CVF Conference on Computer Vision and Pattern Recognition (CVPR)}, 
  title={Objaverse: A Universe of Annotated 3D Objects}, 
  year={2023}
}

@inproceedings{deitke2023objaversexl,
 author = {Deitke, Matt and Liu, Ruoshi and Wallingford, Matthew and Ngo, Huong and Michel, Oscar and Kusupati, Aditya and Fan, Alan and Laforte, Christian and Voleti, Vikram and Gadre, Samir Yitzhak and VanderBilt, Eli and Kembhavi, Aniruddha and Vondrick, Carl and Gkioxari, Georgia and Ehsani, Kiana and Schmidt, Ludwig and Farhadi, Ali},
 booktitle = {Advances in Neural Information Processing Systems (NIPS)},
 title = {Objaverse-XL: A Universe of 10M+ 3D Objects},
 year = {2023}
}

@article{su2024gt23d,
  title={GT23D-Bench: A Comprehensive General Text-to-3D Generation Benchmark},
  author={Su, Sitong and Cai, Xiao and Gao, Lianli and Zeng, Pengpeng and Du, Qinhong and Li, Mengqi and Shen, Heng Tao and Song, Jingkuan},
  journal={arXiv preprint arXiv:2412.09997},
  year={2024}
}

@article{jiang2024survey,
  title={A survey on text-to-3d contents generation in the wild},
  author={Jiang, Chenhan},
  journal={arXiv preprint arXiv:2405.09431},
  year={2024}
}

@inproceedings{hessel2021clipscore,
    title = {{CLIPS}core: A Reference-free Evaluation Metric for Image Captioning},
    author = {Hessel, Jack and Holtzman, Ari and Forbes, Maxwell and Le Bras, Ronan and Choi, Yejin},
    booktitle = {Empirical Methods in Natural Language Processing (EMNLP)},
    year = {2021}
}

@inproceedings{chen2025lara,
  title={Lara: Efficient large-baseline radiance fields},
  author={Chen, Anpei and Xu, Haofei and Esposito, Stefano and Tang, Siyu and Geiger, Andreas},
  booktitle={European Conference on Computer Vision (ECCV)},
  year={2025}
}

@inproceedings{tang2024lgm,
  title={Lgm: Large multi-view gaussian model for high-resolution 3d content creation},
  author={Tang, Jiaxiang and Chen, Zhaoxi and Chen, Xiaokang and Wang, Tengfei and Zeng, Gang and Liu, Ziwei},
  booktitle={European Conference on Computer Vision (ECCV)},
  year={2024}
}

@misc{meshy,
  author       = {AI-5 Meshy},
  title        = {Meshy},
  year         = {2025},
  howpublished = {\url{https://www.meshy.ai/}}
}

@misc{tripo,
  author       = {AI-2.5 Tripo},
  title        = {Tripo},
  year         = {2024},
  howpublished = {\url{https://www.tripo3d.ai/}}
}

@article{wu2023human,
  title={Human preference score v2: A solid benchmark for evaluating human preferences of text-to-image synthesis},
  author={Wu, Xiaoshi and Hao, Yiming and Sun, Keqiang and Chen, Yixiong and Zhu, Feng and Zhao, Rui and Li, Hongsheng},
  journal={arXiv preprint arXiv:2306.09341},
  year={2023}
}

@inproceedings{xu2024imagereward,
  title={Imagereward: Learning and evaluating human preferences for text-to-image generation},
  author={Xu, Jiazheng and Liu, Xiao and Wu, Yuchen and Tong, Yuxuan and Li, Qinkai and Ding, Ming and Tang, Jie and Dong, Yuxiao},
  booktitle={Advances in Neural Information Processing Systems (NIPS)},
  year={2024}
}

@inproceedings{wang2023exploring,
  title={Exploring clip for assessing the look and feel of images},
  author={Wang, Jianyi and Chan, Kelvin CK and Loy, Chen Change},
  booktitle={Proceedings of the AAAI Conference on Artificial Intelligence},
  year={2023}
}

@inproceedings{liu2025clip,
  title={CLIP-PCQA: Exploring Subjective-Aligned Vision-Language Modeling for Point Cloud Quality Assessment},
  author={Liu, Yating and Zhang, Yujie and Shan, Ziyu and Xu, Yiling},
  booktitle={Proceedings of the AAAI Conference on Artificial Intelligence},
  year={2025}
}

@article{huang2026focal,
  title={Focal Reward: Balanced Reinforcement Learning under Rubric-Based Rewards},
  author={Huang, Yu and Zhao, Zihua and Huan, Zhaoxin and Gu, Wanli and Hong, Feng and Ge, Xinmu and Yuan, Lin and Wu, Weichang and Hu, Qiang and Zhang, Xiaolu and others},
  journal={arXiv preprint arXiv:2605.26579},
  year={2026}
}

@article{di2025hyper,
  title={Hyper-3DG: Text-to-3d gaussian generation via hypergraph},
  author={Di, Donglin and Yang, Jiahui and Luo, Chaofan and Xue, Zhou and Chen, Wei and Yang, Xun and Gao, Yue},
  journal={International Journal of Computer Vision (IJCV)},
  year={2025}
}

@article{liu2026advancesgrpogenerationmodels,
  title={Advances in GRPO for Generation Models: A Survey},
  author={Zexiang Liu and Xianglong He and Yangguang Li},
  journal={arXiv preprint arXiv:2603.06623},
  year={2026}
}

@inproceedings{chen2025meshanything,
  title={Meshanything: Artist-created mesh generation with autoregressive transformers},
  author={Chen, Yiwen and He, Tong and Huang, Di and Ye, Weicai and Chen, Sijin and Tang, Jiaxiang and Cai, Zhongang and Yang, Lei and Yu, Gang and Lin, Guosheng and others},
  booktitle={International Conference on Learning Representations (ICLR)},
  year={2025}
}

@inproceedings{siddiqui2024meshgpt,
  title={Meshgpt: Generating triangle meshes with decoder-only transformers},
  author={Siddiqui, Yawar and Alliegro, Antonio and Artemov, Alexey and Tommasi, Tatiana and Sirigatti, Daniele and Rosov, Vladislav and Dai, Angela and Nie{\ss}ner, Matthias},
  booktitle={IEEE/CVF Conference on Computer Vision and Pattern Recognition (CVPR)},
  year={2024}
}

@article{wang2026shapellm,
  title={ShapeLLM-Omni: A Native Multimodal LLM for 3D Generation and Understanding},
  author={Wang, Zhengyi and Zhao, Ruowen and Xie, Shenghao and Zhu, Jun and others},
  journal={Advances in Neural Information Processing Systems (NIPS)},
  year={2026}
}

@inproceedings{zhao2025deepmesh,
  title={Deepmesh: Auto-regressive artist-mesh creation with reinforcement learning},
  author={Zhao, Ruowen and Ye, Junliang and Wang, Zhengyi and Liu, Guangce and Chen, Yiwen and Wang, Yikai and Zhu, Jun},
  booktitle={International Conference on Computer Vision (ICCV)},
  year={2025}
}

@article{wang2024llama,
  title={Llama-mesh: Unifying 3d mesh generation with language models},
  author={Wang, Zhengyi and Lorraine, Jonathan and Wang, Yikai and Su, Hang and Zhu, Jun and Fidler, Sanja and Zeng, Xiaohui},
  journal={arXiv preprint arXiv:2411.09595},
  year={2024}
}

@inproceedings{lin2026meshripple,
  title={Meshripple: Structured autoregressive generation of artist-meshes},
  author={Lin, Junkai and Long, Hang and Guo, Huipeng and Zhang, Jielei and Yang, JiaYi and Guo, Tianle and Yang, Yang and Li, Jianwen and Zhang, Wenxiao and Nie{\ss}ner, Matthias and others},
  booktitle={IEEE/CVF Conference on Computer Vision and Pattern Recognition (CVPR)},
  year={2026}
}

@article{xu2026meshweaver,
  title={MeshWeaver: Sparse-Voxel-Guided Surface Weaving for Autoregressive Mesh Generation},
  author={Xu, Jiale and Zhao, Wang and Shan, Ying},
  journal={arXiv preprint arXiv:2606.04688},
  year={2026}
}

@inproceedings{shen2026flashmesh,
  title={FlashMesh: Faster and Better Autoregressive Mesh Synthesis via Structured Speculation},
  author={Shen, Tingrui and Zhang, Yiheng and Tang, Chen and Ping, Chuan and Zhao, Zixing and Wan, Le and Wang, Yuwang and Wang, Ronggang and He, Shengfeng},
  booktitle={IEEE/CVF Conference on Computer Vision and Pattern Recognition (CVPR)},
  year={2026}
}

@article{wang2026face,
  title={FACE: A Face-based Autoregressive Representation for High-Fidelity and Efficient Mesh Generation},
  author={Wang, Hanxiao and Guo, Yuan-Chen and Liu, Ying-Tian and Zou, Zi-Xin and Zhang, Biao and Quan, Weize and Liang, Ding and Cao, Yan-Pei and Yan, Dong-Ming},
  journal={arXiv preprint arXiv:2603.01515},
  year={2026}
}

@inproceedings{tang2026we,
  title={Are we ready for rl in text-to-3d generation? a progressive investigation},
  author={Tang, Yiwen and Guo, Zoey and Zhu, Kaixin and Zhang, Ray and Chen, Qizhi and Jiang, Dongzhi and Liu, Junli and Zeng, Bohan and Song, Haoming and Qu, Delin and others},
  booktitle={IEEE/CVF Conference on Computer Vision and Pattern Recognition (CVPR)},
  year={2026}
}

@inproceedings{han2026var,
  title={VAR-3D: View-aware Auto-Regressive Model for Text-to-3D Generation via a 3D Tokenizer},
  author={Han, Zongcheng and Hong, Yu and Sun, Haoran and Cao, Dongyan},
  booktitle={International Conference on Multimedia Retrieval (ICMR)},
  year={2026}
}

@article{yi2026progressive,
  title={Progressive text-to-3d generation for automatic 3d prototyping},
  author={Yi, Han and Zheng, Zhedong and Xu, Xiangyu and Chua, Tat-seng},
  journal={ACM Transactions on Multimedia Computing, Communications and Applications (ACM MCCA)},
  year={2026}
}

@article{liu2026hifi3d,
  title={HiFi3D: Improving Text-to-3D With High-Fidelity Multi-View Diffusion},
  author={Liu, Runxin and Chen, Yang and Pan, Yingwei and Xie, Hongtao and Zhang, Yongdong and Yao, Ting and Mei, Tao},
  journal={IEEE Transactions on Multimedia (TMM)},
  year={2026}
}

@article{miao2026dreamer,
  title={Dreamer xl: Towards high-resolution text-to-3d generation via trajectory score matching},
  author={Miao, Xingyu and Duan, Haoran and Ouyang, Zizhou and Hu, Runze and Shah, Tejal and Ojha, Varun and Long, Yang and Ranjan, Rajiv},
  journal={IEEE Transactions on Circuits and Systems for Video Technology (TCSVT)},
  year={2026}
}

@article{zhou2026consdreamer,
  title={ConsDreamer: Advancing Multi-View Consistency for Zero-Shot Text-to-3D Generation},
  author={Zhou, Yuan and Jin, Shilong and Hua, Litao and Lv, Wanjun and Duan, Haoran and Han, Jungong},
  journal={IEEE Transactions on Image Processing (TIP)},
  year={2026},
}

@article{qin2026distilling,
  title={Distilling multi-view diffusion models into 3d generators},
  author={Qin, Hao and Chen, Luyuan and Kong, Ming and Lu, Mengxu and Zhu, Qiang},
  journal={IEEE Transactions on Multimedia (TMM)},
  year={2026}
}

@article{ji2026flux3d,
  title={FLUX3D: High-Fidelity 3D Gaussian Generation with Diffusion-Aligned Sparse Representation},
  author={Ji, Haorui and Liu, Weizhe and Li, Hongdong and Guo, Hengkai},
  journal={arXiv preprint arXiv:2606.24874},
  year={2026}
}

@article{qian2026gs,
  title={GS-Voxel: Fitting-Free Structured Latents for Large-Scale 3DGS Generation},
  author={Qian, Ming and Wang, Zijian and Sun, Minchao and Xiong, Jincheng and Zhang, Hang and Xu, Mu and Wang, Chi and Chen, Baoquan},
  journal={arXiv preprint arXiv:2608.17988},
  year={2026}
}

@article{wu2026animateanymesh++,
  title={AnimateAnyMesh++: A Flexible Feed-Forward Framework for High-Fidelity Text-Driven Mesh Animation},
  author={Wu, Zijie and Yu, Chaohui and Wang, Fan and Bai, Xiang},
  journal={IEEE Transactions on Pattern Analysis and Machine Intelligence (TPAMI)},
  year={2026}
}

@article{lipman2022flow,
  title={Flow matching for generative modeling},
  author={Lipman, Yaron and Chen, Ricky TQ and Ben-Hamu, Heli and Nickel, Maximilian and Le, Matt},
  journal={arXiv preprint arXiv:2210.02747},
  year={2022}
}

@inproceedings{liu2026mracl,
  title={MRACL: Multi-Reward Space Guided Adaptive Curriculum Reinforcement Learning for LLMs},
  author={Liu, Wenxuan and Huo, Liangyu and Jing, Yi and Zhang, Xiyuan and Xie, Jian},
  booktitle={Proceedings of the AAAI Conference on Artificial Intelligence},
  year={2026}
}

@inproceedings{black2024training,
  title={Training diffusion models with reinforcement learning},
  author={Black, Kevin and Janner, Michael and Du, Yilun and Kostrikov, Ilya and Levine, Sergey},
  booktitle={International Conference on Learning Representations (ICLR)},
  year={2024}
}

@article{shao2024deepseekmath,
  title={Deepseekmath: Pushing the limits of mathematical reasoning in open language models},
  author={Shao, Zhihong and Wang, Peiyi and Zhu, Qihao and Xu, Runxin and Song, Junxiao and Bi, Xiao and Zhang, Haowei and Zhang, Mingchuan and Li, YK and Wu, Yang and others},
  journal={arXiv preprint arXiv:2402.03300},
  year={2024}
}

@inproceedings{wallace2024diffusion,
  title={Diffusion model alignment using direct preference optimization},
  author={Wallace, Bram and Dang, Meihua and Rafailov, Rafael and Zhou, Linqi and Lou, Aaron and Purushwalkam, Senthil and Ermon, Stefano and Xiong, Caiming and Joty, Shafiq and Naik, Nikhil},
  booktitle={IEEE/CVF Conference on Computer Vision and Pattern Recognition (CVPR)},
  year={2024}
}

@article{liu2026improving,
  title={Improving video generation with human feedback},
  author={Liu, Jie and Liu, Gongye and Liang, Jiajun and Yuan, Ziyang and Liu, Xiaokun and Zheng, Mingwu and Wu, Xiele and Wang, Qiulin and Xia, Menghan and Wang, Xintao and others},
  journal={Advances in Neural Information Processing Systems (NIPS)},
  year={2026}
}

@inproceedings{ding2026treegrpo,
  title={Treegrpo: Tree-advantage grpo for online rl post-training of diffusion models},
  author={Ding, Zheng and Ye, Weirui},
  booktitle={International Conference on Learning Representations (ICLR)},
  year={2026}
}

@inproceedings{rojas2026improving,
  title={Improving reasoning for diffusion language models via group diffusion policy optimization},
  author={Rojas, Kevin and Lin, Jiahe and Rasul, Kashif and Schneider, Anderson and Nevmyvaka, Yuriy and Tao, Molei and Deng, Wei},
  booktitle={International Conference on Learning Representations (ICLR)},
  year={2026}
}

@inproceedings{luo2026reinforcing,
  title={Reinforcing diffusion models by direct group preference optimization},
  author={Luo, Yihong and Hu, Tianyang and Tang, Jing},
  booktitle={International Conference on Learning Representations (ICLR)},
  year={2026}
}

@article{peng2026self,
  title={Self-supervised direct preference optimization for text-to-image diffusion models},
  author={Peng, Liang and Wu, Boxi and Cheng, Haoran and Zhao, Yibo and He, Xiaofei},
  journal={Advances in Neural Information Processing Systems (NIPS)},
  year={2026}
}

@article{lyu2025multi,
  title={Multi-GRPO: Multi-Group Advantage Estimation for Text-to-Image Generation with Tree-Based Trajectories and Multiple Rewards},
  author={Lyu, Qiang and Chen, Zicong and Wang, Chongxiao and Shi, Haolin and Gao, Shibo and Piao, Ran and Zeng, Youwei and Si, Jianlou and Ding, Fei and Li, Jing and others},
  journal={arXiv preprint arXiv:2512.00743},
  year={2025}
}

@article{zhao2026marble,
  title={MARBLE: Multi-Aspect Reward Balance for Diffusion RL},
  author={Zhao, Canyu and Chen, Hao and Tong, Yunze and Qiao, Yu and Li, Jiacheng and Shen, Chunhua},
  journal={arXiv preprint arXiv:2605.06507},
  year={2026}
}

@article{zhang2026aligning,
  title={Aligning few-step diffusion models with dense reward difference learning},
  author={Zhang, Ziyi and Shen, Li and Zhang, Sen and Ye, Deheng and Luo, Yong and Shi, Miaojing and Shan, Dongjing and Du, Bo and Tao, Dacheng},
  journal={IEEE Transactions on Pattern Analysis and Machine Intelligence (TPAMI)},
  year={2026}
}

@inproceedings{lee2024parrot,
  title={Parrot: Pareto-optimal multi-reward reinforcement learning framework for text-to-image generation},
  author={Lee, Seung Hyun and Li, Yinxiao and Ke, Junjie and Yoo, Innfarn and Zhang, Han and Yu, Jiahui and Wang, Qifei and Deng, Fei and Entis, Glenn and He, Junfeng and others},
  booktitle={European Conference on Computer Vision (ECCV)},
  year={2024}
}

@article{jiang2026t2i,
  title={T2i-r1: Reinforcing image generation with collaborative semantic-level and token-level cot},
  author={Jiang, Dongzhi and Guo, Ziyu and Zhang, Renrui and Zong, Zhuofan and Li, Hao and Zhuo, Le and Yan, Shilin and Heng, Pheng-Ann and Li, Hongsheng},
  journal={Advances in Neural Information Processing Systems (NIPS)},
  year={2026}
}

@article{liu2021conflict,
  title={Conflict-averse gradient descent for multi-task learning},
  author={Liu, Bo and Liu, Xingchao and Jin, Xiaojie and Stone, Peter and Liu, Qiang},
  journal={Advances in neural information processing systems (NIPS)}}

@article{rame2023rewarded,
  title={Rewarded soups: towards pareto-optimal alignment by interpolating weights fine-tuned on diverse rewards},
  author={Rame, Alexandre and Couairon, Guillaume and Dancette, Corentin and Gaya, Jean-Baptiste and Shukor, Mustafa and Soulier, Laure and Cord, Matthieu},
  journal={Advances in Neural Information Processing Systems (NIPS)},
  year={2023}
}

@article{sener2018multi,
  title={Multi-task learning as multi-objective optimization},
  author={Sener, Ozan and Koltun, Vladlen},
  journal={Advances in neural information processing systems (NIPS)},
  year={2018}
}

@article{yu2020gradient,
  title={Gradient surgery for multi-task learning},
  author={Yu, Tianhe and Kumar, Saurabh and Gupta, Abhishek and Levine, Sergey and Hausman, Karol and Finn, Chelsea},
  journal={Advances in neural information processing systems (NIPS)},
  year={2020}
}

@article{desideri2012multiple,
  title={Multiple-gradient descent algorithm (MGDA) for multiobjective optimization},
  author={D{\'e}sid{\'e}ri, Jean-Antoine},
  journal={Comptes Rendus. Math{\'e}matique},
  year={2012}
}

@inproceedings{lai2026lattice,
  title={Lattice: Democratize high-fidelity 3d generation at scale},
  author={Lai, Zeqiang and Zhao, Yunfei and Zhao, Zibo and Liu, Haolin and Lin, Qingxiang and Huang, Jingwei and Guo, Chunchao and Yue, Xiangyu},
  booktitle={{IEEE/CVF Conference on Computer Vision and Pattern Recognition (CVPR)}},
  year={2026}
}

@inproceedings{zhou2024uni3d,
  title={Uni3d: Exploring unified 3d representation at scale},
  author={Zhou, Junsheng and Wang, Jinsheng and Ma, Baorui and Liu, Yu-Shen and Huang, Tiejun and Wang, Xinlong},
  booktitle={International Conference on Learning Representations (ICLR)},
  year={2024}
}

@article{li2025triposg,
  title={Triposg: High-fidelity 3d shape synthesis using large-scale rectified flow models},
  author={Li, Yangguang and Zou, Zi-Xin and Liu, Zexiang and Wang, Dehu and Liang, Yuan and Yu, Zhipeng and Liu, Xingchao and Guo, Yuan-Chen and Liang, Ding and Ouyang, Wanli and others},
  journal={IEEE Transactions on Pattern Analysis and Machine Intelligence (TPAMI)},
  year={2025}
}

@inproceedings{xue2023ulip,
  title={Ulip: Learning a unified representation of language, images, and point clouds for 3d understanding},
  author={Xue, Le and Gao, Mingfei and Xing, Chen and Mart{\'\i}n-Mart{\'\i}n, Roberto and Wu, Jiajun and Xiong, Caiming and Xu, Ran and Niebles, Juan Carlos and Savarese, Silvio},
  booktitle={IEEE/CVF Conference on Computer Vision and Pattern Recognition (CVPR)},
  year={2023},
}

@article{sun2026meshflow,
  title={MeshFlow: Mesh Generation with Equivariant Flow Matching},
  author={Sun, Qi and Nakayama, Kiyohiro and Yan, Jing Nathan and Huang, Qixing and Rush, Alexander and Guibas, Leonidas and Wetzstein, Gordon and Liao, Jing and Yang, Guandao},
  journal={arXiv preprint arXiv:2606.23489},
  year={2026}
}

@article{xu2026meshyt2,
  title={Meshy T2: Fast Native Mesh Generation with Flow Matching},
  author={Xu, Jiale and Liang, Rendong and Long, Yuhao and Shen, Siyuan and Xian, Zangyueyang and Xu, Zeyi and Hu, Yuanming},
  journal={arXiv preprint arXiv:2607.28675},
  year={2026}
}

@article{fei2026home3d,
  title={Home3D 1.0: A High-Fidelity Image-to-3D Asset Generation System for Interior Design},
  author={Fei, Yiyun and Li, Guoqiu and Song, Jin and Wu, Chuqiao and Wu, Delong and Wu, Hong and Zeng, Ziru and Chen, Haohui and Kong, Yindong and Li, Jing and others},
  journal={arXiv preprint arXiv:2606.27923},
  year={2026}
}

@article{li2024hunyuan,
  title={Hunyuan-dit: A powerful multi-resolution diffusion transformer with fine-grained chinese understanding},
  author={Li, Zhimin and Zhang, Jianwei and Lin, Qin and Xiong, Jiangfeng and Long, Yanxin and Deng, Xinchi and Zhang, Yingfang and Liu, Xingchao and Huang, Minbin and Xiao, Zedong and others},
  journal={arXiv preprint arXiv:2405.08748},
  year={2024}
}
\end{document}